\documentclass[journal,twoside]{IEEEtran}

\usepackage{xcolor}

\makeatletter
\def\subsectioncolor#1{}
\def\sectioncolor#1{}
\def\IEEElogo#1{}
\def\IEEEsetpublogo#1{}
\makeatother

\usepackage{cite}
\usepackage{amsmath,amssymb,amsfonts}
\usepackage{algorithmic}
\usepackage{graphicx}
\usepackage{hyperref}
\hypersetup{hidelinks=true}
\usepackage{textcomp}
\usepackage{cite}
\usepackage[switch]{lineno}
\usepackage{amsmath,amssymb,amsfonts}
\usepackage{graphicx}
\usepackage{textcomp}
\usepackage{framed,multirow}
\usepackage{booktabs}
\usepackage{multirow}
\usepackage{amsmath}
\usepackage{array}
\usepackage{verbatim}
\usepackage{amsmath}
\usepackage{amssymb}
\usepackage{xcolor}
\usepackage{multirow} 
\usepackage{tabularx}
\usepackage{colortbl}
\usepackage{xcolor}
\usepackage{array}
\usepackage{makecell}
\usepackage[ruled,linesnumbered]{algorithm2e}
\newcolumntype{C}[1]{>{\centering\arraybackslash}p{#1}}
\usepackage{amssymb}
\usepackage{latexsym}
\usepackage{float}
\usepackage{array}
\usepackage{enumitem}
\usepackage{caption}
\usepackage{tabularx}
\usepackage{booktabs}
\usepackage{pifont}
\usepackage{tabularx}
\usepackage{stfloats}
\usepackage{url}
\usepackage{xcolor}
\usepackage{hyperref}
\begin{document}
\title{Pattern-Aware Diffusion Synthesis of fMRI/dMRI with Tissue and Microstructural Refinement}
\author{Xiongri Shen,  Jiaqi Wang, Yi Zhong, Zhenxi Song, \textit{Member},Leilei Zhao, Yichen Wei, Lingyan Liang, Shuqiang Wang, \textit{Senior Member}, Baiying Lei, \textit{Senior Member}, Demao Deng, Zhiguo Zhang, \textit{Member}
%\vspace{-11mm}
\thanks{
This research was supported by the National Natural Science Foundation of China (Grants 62306089, 32361143787, 82102032), the China Postdoctoral Science Foundation (Grants 2023M730873, GZB20230960), the key Project of Basic Research of Shenzhen (NO: JCYJ20200109113603854), and the Guangxi Natural Science Foundation (Grant No. 2023GXNS-FBA026073), the Shenzhen Science and Technology Program(Grant No. RCBS20231211090800003) (Corresponding authors: Zhenxi Song and Zhiguo Zhang.)
%This paragraph of the first footnote will contain the date on which
%you submitted your paper for review. It will also contain support information,
%including sponsor and financial support acknowledgment. For example, 
%``This work was supported in part by the U.S. Department of Commerce under Grant BS123456.'' 
}
%\thanks{The next few paragraphs should contain the authors' current affiliations, including current address and e-mail. 
% For example, F. A. Author is with the National Institute of Standards and Technology, Boulder, CO 80305 USA (e-mail:author@boulder.nist.gov). }
\thanks{Xiongri Shen, Yi Zhong, Jiaqi Wang,  and Leilei Zhao  are with the Department of Computer Science and Technology, Harbin Institute of Technology, Shenzhen, 518055, China (email: xiongrishen@stu.hit.edu.cn, zoey24@stu.hit.edu.cn, 23b951063@stu.hit.edu.cn, 24b951025@stu.hit.edu.cn ).}
% \thanks{
% Linling Li is with the Guangdong Key Laboratory of Biomedical Measurements and Ultrasound Imaging, School of Biomedical Engineering, Shenzhen University Medical School, Shenzhen University, Shenzhen 518060, China (lilinling@szu.edu.cn).}
\thanks{
Zhenxi Song and Zhiguo Zhang is with School of Intelligence Science and Engineering, College of Artificial Intelligence, Harbin Institute of Technology, Shenzhen, Guangdong, China(e-mail: songzhenxi@hit.edu.cn, zhiguozhang@hit.edu.cn).
}
\thanks{
Baiying Lei is with School of Biomedical Engineering, National-Regional Key Technology Engineering Laboratory for Medical Ultrasound, Guangdong Key Laboratory for Biomedical, Measurements and Ultrasound Imaging, Shenzhen University Medical School, Shenzhen University, Shenzhen, China (e-mail: leiby@szu.edu.cn).
}
\thanks{
Shuqiang Wang is with  the Shenzhen Institutes of
 Advanced Technology, Chinese Academy of Sciences, Shenzhen, China
 (e-mail: sq.wang@siat.ac.cn).
}
\thanks{
Demao Deng, Yichen Wei, and Lingyan Liang are with the Department of Radiology, The People’s Hospital of Guangxi Zhuang Autonomous Region, Guangxi Academy of Medical Sciences. Nanning, China (email: demaodeng@163.com, 316644690@qq.com, lianglingyan163@126.com).}
}

\maketitle

\begin{abstract}
Magnetic resonance imaging (MRI), especially functional MRI (fMRI) and diffusion MRI (dMRI), is essential for studying neurodegenerative diseases. However, missing modalities pose a major barrier to their clinical use. Although GAN- and diffusion model-based approaches have shown some promise in modality completion, they remain limited in fMRI-dMRI synthesis due to (1) significant BOLD vs. diffusion-weighted signal differences between fMRI and dMRI in time/gradient axis, and (2) inadequate integration of disease-related neuroanatomical patterns during generation. To address these challenges, we propose PDS, introducing two key innovations: (1) a pattern-aware dual-modal 3D diffusion framework for cross-modality learning, and (2) a tissue refinement network integrated with a efficient microstructure refinement to maintain structural fidelity and fine details. Evaluated on OASIS-3, ADNI, and in-house datasets, our method achieves state-of-the-art results, with PSNR/SSIM scores of 29.83 dB/90.84\% for fMRI synthesis (+1.54 dB/+4.12\% over baselines) and 30.00 dB/77.55\% for dMRI synthesis (+1.02 dB/+2.2\%). In clinical validation, the synthesized data show strong diagnostic performance, achieving 67.92\%/66.02\%/64.15\% accuracy (NC vs. MCI vs. AD) in hybrid real-synthetic experiments. Code is available in \href{https://github.com/SXR3015/PDS}{PDS GitHub Repository}
\end{abstract}


\section{Introduction}
\label{sec:introduction}
\IEEEPARstart{M}{agnetic} Resonance Imaging (MRI) has transformed medical diagnosis by enabling non-invasive visualization of anatomical structures with high soft tissue contrast. Two specialized MRI techniques have further advanced neuroimaging: functional MRI (fMRI) captures blood oxygen level-dependent (BOLD) signals to measure brain functions, while diffusion magnetic resonance imaging (dMRI) maps white matter integrity through water diffusion \cite{irnich1995magnetostimulation,moseley1995clinical,lim2002neuropsychiatric,assaf2008diffusion,smith2004overview}. However, incompatible echo planar imaging (EPI) sequences of fMRI and dMRI prevent their concurrent acquisition, often resulting in missing modalities in clinical practice. This sequential MRI scanning approach increases costs, examination duration, and motion artifacts — limitations particularly challenging in resource-constrained settings. Cross-modal synthesis offers a promising solution by computationally generating missing MRI modalities from existing ones, enabling complete diagnostic information while addressing acquisition constraints \cite{nie2018medical,yu2020medical}. \textbf{fMRI and dMRI are 4D data with time/gradient axis, however, they lack of the implicit relationship in time/gradient axis}. Therefore, using 3D cross-modal synthesis approach for mean data is more  robust and resonable. 3D data is also critical for computing imaging biomarkers (e.g., ROI signals) \cite{qin2025amplitude, chou2023functional}. 
 % Clinical MRI often lacks certain modalities due to cost, safety, and time constraints, especially between sequences with different EPI protocols (e.g., fMRI vs. dMRI) \cite{chou2023functional}. 
 While existing synthesis methods based on the U-Net family (UF), they fail to bridge fundamental 3D MRI modality gaps \cite{son2019synthesizing, ren2021q}. Specifically, current approaches overlook divergence between fMRI and dMRI, and ignore disease semantics. 

%Therefore, cross-modal MRI synthesis responds to the growing needs for efficient and cost-effective neuroimaging protocols.

%  limitaion of current generative methods  
    U-Net generates images through direct transformation, but is limited by its representation capabilities. Generative adversarial networks (GANs) and diffusion models (DMs) have emerged as two leading approaches for medical image synthesis, with GANs leveraging adversarial training and DMs employing iterative denoising to generate medical images (including neuroimages)  \cite{goodfellow2020generative,croitoru2023diffusion,yang2023diffusion}. GAN variants, such as self-attention GAN (SA-GAN) \cite{zhang2019self}, perceptual GAN (pGAN) \cite{liu2019perceptual}, etc., establish cross-modal mappings via adversarial training between generators and discriminators. While effective for 2D data, GAN variants struggle with 3D cross-modal MRI synthesis, often experiencing model collapse due to the  heterogeneity and inherent sparsity between 3D source and target domains (fMRI or dMRI ) \cite{liu2021ct,tavse2022systematic,islam2020gan}. 
   
     \begin{figure*}[ht]
    \centering
    \includegraphics[width=0.96\textwidth]{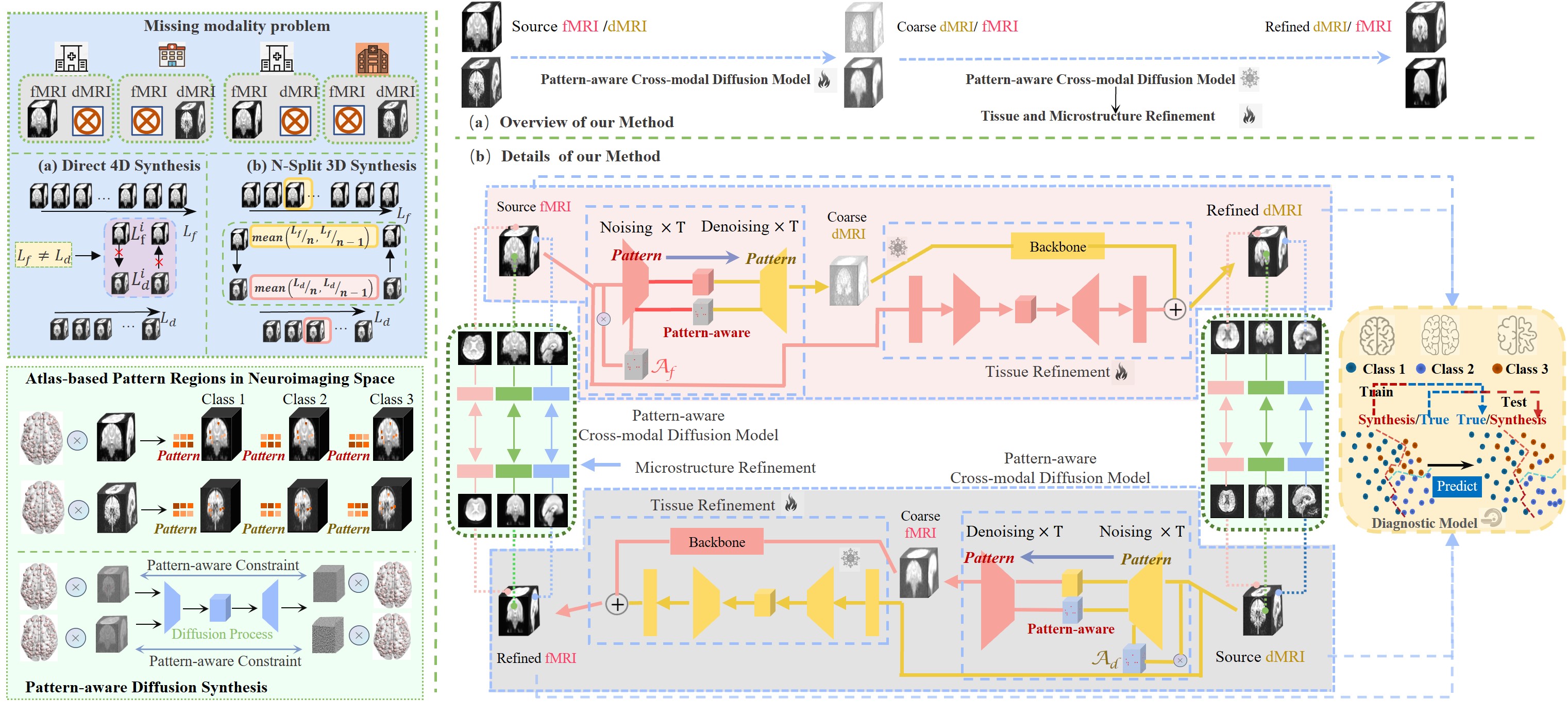} % Insert image, width set to full text width
    \vspace{-3mm}
    \caption{
We divide the uneven time axis into N segments to align the temporal/gradient dimensions between fMRI and dMRI. Then, to implement the cross-modal fMRI/DTI synthesis, we employ the pattern-aware dual-modal diffusion model (PDM), tissue refinement (TR), and microstructure refinement (MR) to improve the image quality and clinical applicability. } % Caption for the image
    
    \label{fig:0} % Label for referencing the image
        \vspace{-6mm}

\end{figure*}
   
   DMs employ a unique generation paradigm that reconstructs data through Markov chain-based iterative denoising of corrupted inputs. Although it shows a stronger theoretical robustness than GAN in 2D image synthesis, their noise estimation strategy does not scale effectively for 3D volumes that exhibit amplified fMRI/dMRI heterogeneity \cite{pan20232d,pan2024full,jiang2023cola}, leading to a poorer performance in 3D neuroimage synthesis compared to GAN-based architectures in some scenario 
   (3D fMRI/dMRI synthesis, e.g.) 
   \cite{lei2024diffusiongan3d}.
   
   However, two key challenges remain for both GAN- and DM-based fMRI/dMRI synthesis approaches.     
    (1) \textbf{BOLD vs. diffusion-weighted signal difference}: it is challenging to bridge the substantial discrepancy between fMRI-derived BOLD and dMRI's microstructural diffusion properties \cite{mohammadi2024mri}. This multidimensional challenge encompasses both spatial discordance (voxel-level heterogeneity) and functional mismatch (neurophysiological-process decoupling), resulting in compromised semantic accuracy, tissue specificity, and microstructural fidelity in synthesized neuroimages. (2) \textbf{Semantic dissociation}: current fMRI/dMRI synthesis approaches primarily focus on image-level fidelity while overlooking disease-specific semantics and diagnostic relevance \cite{zhang2022predicting}. This oversight means that synthesized neuroimages, while visually plausible, may not capture subtle pathological features critical for disease diagnosis. Consequently, the clinical utility of these synthetic images remains limited, as they may fail to reflect the distinctive neural signatures of various conditions.

   % To address these challenges, we first integrate disease-specific neuroanatomical patterns (termed \textbf{\emph{disease semantics}}) into neuroimaging synthesis, leveraging anatomically parcellated atlases to identify multimodal neural markers in fMRI/dMRI data. These neural markers exhibit distinct coupling in normal controls (NC) versus decoupling in mild cognitive impairment (MCI) /Alzheimer's disease (AD) due to disrupted functional-structural relationships — a critical feature ignored by conventional 3D diffusion models. To address this, we embed disease semantics as denoising priors, conditioning diffusion steps on atlas-derived functional-structural correlations from NC/MCI/AD cohorts. This ensures synthetic data align with pathological ground truths, bridging fMRI-dMRI discrepancies in 3D space. Further, we employ tissue and microstructure refinement modules to reduce spatial blurring and enhance biophysical consistency. The resulting pattern-aware diffusion model with tissue and microstructure refinements (\textbf{PDS}) framework can resolve 3D cross-modal synthesis limitations while preserving diagnostic relevance. normal controls (NC) and decoupling in mild cognitive impairment (MCI) /Alzheimer's disease (AD)
   
   To address these challenges, we incorporate \textbf{\emph{disease semantics (patterns)}} into neuroimaging synthesis, using anatomical atlases to identify multimodal neural markers in fMRI/dMRI. 
   These markers show coupling in cognitive impair disease due to disrupted functional-structural relationships, a key factor overlooked by conventional 3D diffusion models.
   We embed disease semantics as denoising priors, conditioning diffusion steps on atlas-derived correlations from NC/MCI/AD cohorts, ensuring synthetic data reflect pathological truths and bridge fMRI-dMRI discrepancies. Additionally, tissue and microstructure refinement modules reduce spatial blurring and improve biophysical consistency. The resulting \textbf{P}attern-aware \textbf{D}iffusion \textbf{S}ynthesis (\textbf{PDS}) overcomes 3D cross-modal synthesis limitations while preserving diagnostic relevance. Our key contributions include:
   \begin{itemize}
    \item \textbf{Pattern-aware Dual-modal 3D Diffusion Model.} 
    %R3 diffuser pioneers a dual-branch diffusion framework that synergistically leverages inter-modal dependencies during noise estimation. 
We condition disease semantic estimation during the denoising steps of a dual-modal diffusion model on paired 3D source-target modalities (e.g., fMRI→dMRI as well as dMRI→fMRI). This directly mitigates the 3D cross-modal domain gap that causes feature misalignment in conventional diffusion models, enhancing disease semantic details in the generated fMRI and dMRI. 
    % \item \textbf{tissue and Microstructure Refinement.} To enhance the tissue and microstructure details of synthesis fMRI and dMRI, we introduce tissue and microstructure refinement.
    % \begin{itemize}
    %   \item \textbf{Tissue Refinement.} An tissue refinement network addresses the spatial blurring inherent in diffusion models. Its tissue-projection branch explicitly optimizes microstructural fidelity (e.g., white matter tract geometry in DTI) through learnable tissue priors—overcoming the limitation of pure noise-to-image mappings in diffusion models. 
    %   \item \textbf{ Microstructure Refinement.} We compute microstructure discrepancies in deep feature space between triplanar (axial/coronal/sagittal) projections of target and synthesized fMRI/DWI-DTI images, enabling computationally-efficient 3D microstructure refinement.
    %   \end{itemize}
    % \item \textbf{Clinical applicability.} Except for validating the synthesis fMRI and DWI/DTI by quantitative and visualize comparison on three datasets (OASIS-3, ADNI, and a clinical cohort), we also conduct the diagnostic experiments on the three dataset to validate the clinical applicability of synthesis data.

\item \textbf{Tissue and Efficient Microstructure Refinements.} We introduce (1) \textit{Tissue Refinement}: A projection network with learnable neuroanatomical priors that mitigates spatial blurring caused by noise estimation in the dual-modal diffusion model; (2) \textit{Efficent Microstructure Refinement}: A parameter-efficient module that computes and reduces microstructure-perception feature discrepancies between synthesized and real neuroimaging across axial/coronal/sagittal planes.

\item \textbf{Comprehensive Experiments.} Our synthetic fMRI and dMRI achieve SOTA 3D image fidelity (SSIM=0.908/0.902/0.923 on three datasets: OASIS-3/ADNI/self-collected) while demonstrating clinical efficacy through improved cognitive decline disease classification accuracy (67.92\%/66.02\%/64.15\%), enhanced inter-class discrimination in dual-modal feature space and higher-quality ROI signals. 
\end{itemize}

\section{Related Work}
  % \textbf{MRI Synthesis and Broader Medical Image Synthesis.} Clinical MRI often encounters the problem of missing imaging modality due to constrains in cost, safety, and acquisition time constraints, particularly between MRI sequences with different EPI protocols (e.g., fMRI vs. dMRI) \cite{chou2023functional}. Although current medical image synthesis paradigms focus primarily on reducing radiation doses (e.g., computed tomography to positron emission tomography \cite{dayarathna2024deep}) or standardization of the protocol (T1 to T2 MRI) \cite{kawahara2021t1}, they fail to address the fundamental challenges of bridging distinct 3D MRI modalities. More precisely, spatiotemporal divergence between fMRI and dMRI, and overlooking of disease-semantics are two key and fundamental limitations of existing MRI synthesis methods. Our work addresses this gap by developing a disease-semantic-informed 3D diffusion model and tissue-microstructure refinements. The proposed model is based on DMs and is compared with GANs, so we will first introduce related works using GANs and DMs.
 \textbf{fMRI/dMRI Synthesis.}
Clinical MRI often lacks certain modalities due to cost, safety, and time constraints, especially between sequences with different EPI protocols (e.g., fMRI vs. dMRI) \cite{chou2023functional}. While existing synthesis methods based on the UF, they fail to bridge fundamental 3D MRI modality gaps \cite{son2019synthesizing, ren2021q}. Specifically, current approaches overlook divergence between fMRI and dMRI, and ignore disease semantics.

\textbf{Generative Methods in MRI and Medical Image Synthesis.}
Three major generative architectures in medical image synthesis are the U-Net family (UF) \cite{kalluvila2022synthetic}, GAN family (DF) \cite{xue2023pet,zhan2021multi,zhan2022d2fe}, and diffusion model family (DMF) \cite{peng20232d,zhu2023make,dorjsembe2024conditional,meng2024multi}.  
UF methods generate images via direct transformation but are limited in representational capacity.  
DF methods, including CycleGAN \cite{mahboubisarighieh2024assessing}, cGAN \cite{zhan2021lr}, and DCGAN \cite{luo2024mask}, suffer from instability and mode collapse when bridging large modality gaps or handling sparse 3D MRI data.  

DMF approaches, such as DDPM \cite{chang2024high}, DDIM \cite{ramanarayanan2024dce}, LDM \cite{friedrich2025high}, and DiT \cite{hatamizadeh2024diffit}, iteratively denoise images and show superior generation quality over GANs. However, their step-wise noise estimation is computationally expensive. While LDM improves efficiency via latent-space compression and attention mechanisms, and DiT uses transformers for noise estimation, both struggle with high-quality synthesis due to indirect noise modeling.

Hybrid methods like adversarial diffusion models (ADM) \cite{ozbey2023unsupervised} combine DMs for generation and GANs for refinement. Yet, their reliance on adversarial training makes them prone to instability when translating between 3D fMRI and dMRI.

  \textbf{Disease Semantics.}
MRI captures disease-relevant regions that serve as interpretable neural markers for diagnostic tasks \cite{dosenbach2025brain}. Clinical research demonstrates that functional connectivity in fMRI and structural connectivity in dMRI exhibit distinct \textbf{\emph{patterns}} that evolve with disease progression. For example, normal people show robust and strong activations and connections in fMRI and dMRI, but, in cognitive impairment conditions such as MCI and AD, functional connectivity in fMRI typically strengthens but structural connectivity in dMRI weakens \cite{li2025revealing, ibrahim2021diagnostic}. We define these characteristic \textbf{\emph{patterns}} as \textbf{\emph{disease-semantics}} of the brain. 

\section{Methods}
%Please follow the steps outlined below when submitting your manuscript to
%the IEEE Computer Society Press.  This style guide now has several
%important modifications (for example, you are no longer warned against the
%use of sticky tape to attach your artwork to the paper), so all authors
%should read this new version.
  % To refine the semantic, tissue, and microstructure details of generated neuroimaging, we propose the PDS. Specifically, PDS consists of two blocks: (1)  disease semantic-informed dual-modal 3D diffusion model (DSD3DM) (2) tissue refinement network and microstructure refinement. These two blocks refine the semantic, tissue, and microstructure details of generated images derived from the dual-modal diffusion model.
  %The overall architecture of the PDS is shown in Fig. \ref{fig:1}. 
%-------------------------------------------------------------------------
\begin{figure*}[htbp]
    \centering
    \includegraphics[width=0.95\textwidth]{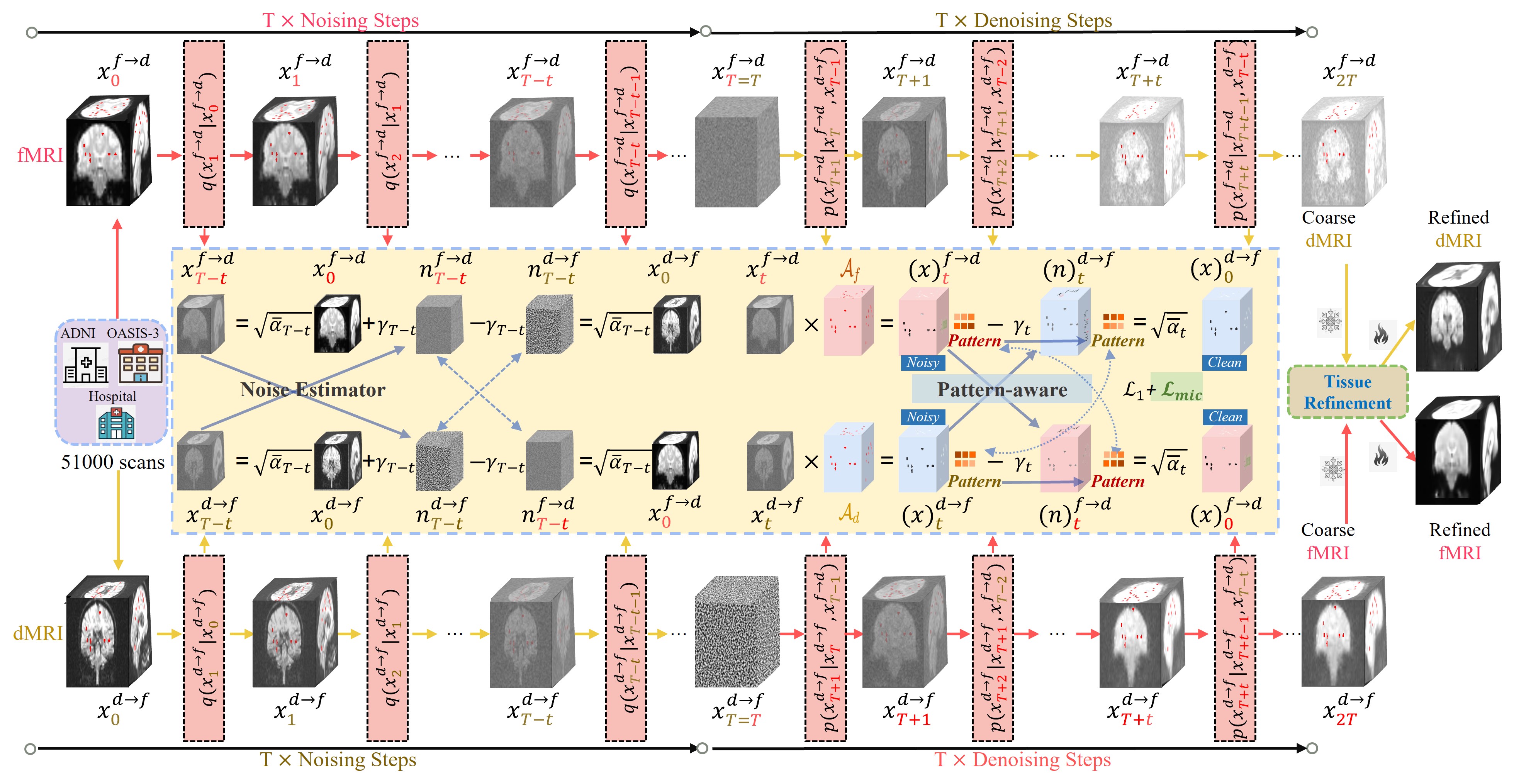} % Insert image, width set to full text width
    \vspace{-3mm}
    \caption{
    Pattern-aware Diffusion Model. Our diffusion model incorporates a noise estimator (NE) for initial fMRI and dMRI generation through progressive denoising. A pattern-aware (DA) module ensures semantic consistency  (pattern at the brain disease-related regions of atlas ) across noisy image and noise. Tissue and microstructure details are refined through a dedicated projection network and a microstructure refinement loss $\mathcal{L}_{mic}$, where $\gamma_t = \sqrt{1 - \bar{\alpha}_t}$. 
    } % Caption for the image
    \label{fig:1} % Label for referencing the image
    \vspace{-3mm}
\end{figure*}

\subsection{pattern-aware Diffusion Model (PDM)}

%refine the fMRI and dMRI images' semantics by semantic-aware dual-modal diffusion model
% We first employ a dual-modal diffusion model to generate baseline fMRI and dMRI images, which are subsequently enhanced through disease-semantics incorporation and refinements. Let $x^{f \rightarrow d}_{0}$ and $x^{d \rightarrow f}_{0}$ represent source fMRI and dMRI images, respectively. For clarity in expressing the tissue refinement network, we define a total time step as $2T$, $0 \leq t \leq T $ denotes the adding noise process, $T < t \leq 2T $ denotes the denoising process. The core innovation of our dual-modal diffusion model lies in estimating 3D noise of the target modality based on 3D noise of source modality. The details of disease-semantic-informed dual-modal 3D diffusion modal are shown in Fig. \ref{fig:1}. The complete process of cross-modal diffusion model generation is as follows:

We employ a dual-modal diffusion model to generate baseline fMRI and dMRI images, which are refined through disease semantics. Let $x^{f \rightarrow d}_{0}$ and $x^{d \rightarrow f}_{0}$ denote the source fMRI and dMRI images. To clarify the tissue refinement network, we define a total time step of $2T$, where $0 \leq t \leq T$ corresponds to noise addition and $T < t \leq 2T$ to denoising. Our key innovation lies in estimating 3D noise of the target modality using the source modality. Details are shown in Fig.~\ref{fig:1}. The full cross-modal generation process is as follows:

\textbf{Dual-modal forward noising process.} The forward noising process \( q \) and denoising process \( p \)  for a given image $x^{f \rightarrow d}_{0}$ and $x^{d \rightarrow f}_{0}$ at step $0 \leq t \leq T$ is given by:
%  {-2mm}
\begin{equation}
   % \[
q(x^{d \rightarrow f}_t | x^{d \rightarrow f}_{t-1}) := \mathcal{N}(x^{d \rightarrow f}_t; \sqrt{1 - \beta_t} x^{d \rightarrow f}_{t-1}, \beta_t \boldsymbol{I}),
%\]
\label{eq:1}
\end{equation}
\begin{equation}
   % \[
q(x^{f \rightarrow d}_t | x^{f \rightarrow d}_{t-1}) := \mathcal{N}(x^{f \rightarrow d}_t; \sqrt{1 - \beta_t} x^{f \rightarrow d}_{t-1}, \beta_t \boldsymbol{I}),
%\]
\label{eq:2}
\end{equation}
where \( \boldsymbol{I} \) denotes the identity matrix, \( \beta_1, \ldots, \beta_T \) are the forward process variances, $ \{x^{f \rightarrow d}_T, x^{d \rightarrow f}_T\} \sim \mathcal{N} (0, \boldsymbol{I})$.

By applying the forward process over \(t\) steps in both modal branches, Equations \ref{eq:1} and \ref{eq:2} can be generalized as:
\begin{equation}
q(x^{d \rightarrow f}_t | x^{d \rightarrow f}_0) := \mathcal{N}(x^{d \rightarrow f}_t; \sqrt{\bar{\alpha}_t} x^{d \rightarrow f}_0, (1 - \bar{\alpha}_t) \boldsymbol{I}),
\end{equation}
\begin{equation}
q(x^{f \rightarrow d}_t | x^{f \rightarrow d}_0) := \mathcal{N}(x^{f \rightarrow d}_t; \sqrt{\bar{\alpha}_t} x^{f \rightarrow d}_0, (1 - \bar{\alpha}_t) \boldsymbol{I}),
\end{equation}
with \( \alpha_t := 1 - \beta_t \) and \( \bar{\alpha}_t := \prod_{s=1}^t \alpha_s \).

\( x^{d \rightarrow f}_t \) and \( x^{f \rightarrow d}_t \) can be
expressed via the re-parameterization trick as
:
\begin{equation}
x^{d \rightarrow f}_t = \sqrt{\bar{\alpha}_t} x^{d \rightarrow f}_0 + \sqrt{1 - \bar{\alpha}_t} \epsilon, \quad \text{with } \epsilon \sim \mathcal{N}(0, \boldsymbol{I}),
\end{equation}
\begin{equation}
x^{f \rightarrow d}_t = \sqrt{\bar{\alpha}_t} x^{f \rightarrow d}_0 + \sqrt{1 - \bar{\alpha}_t} \epsilon, \quad \text{with } \epsilon \sim \mathcal{N}(0, \boldsymbol{I}).
\end{equation}

\textbf{Cross-modal inverse denoising process.} \( x^{d \rightarrow f}_{T+t} \) and \( x^{f \rightarrow d}_{T+t} \) are predicted based on \( x^{f \rightarrow d}_{T-t} \) and \( x^{d \rightarrow f}_{T-t} \). Unlike conventional denoising processes, our dual-modal diffusion model leverages noisy images from complementary modalities for noise estimation. A cross-modal noise estimator generates the initial fMRI and dMRI outputs.

\begin{equation}
x^{d \rightarrow f}_{T+t+1} = \frac{1}{\sqrt{\alpha_t}} \left( x^{d \rightarrow f}_{T+t} - \frac{1 - \alpha_t}{\sqrt{1 - \bar{\alpha}_t}} \epsilon_\theta( x^{\boldsymbol{f \rightarrow d}}_{t}, t) \right) + \sigma_t \boldsymbol{z},
\end{equation}
\begin{equation}
x^{f \rightarrow d}_{T+t+1} = \frac{1}{\sqrt{\alpha_t}} \left( x^{f \rightarrow d}_{T+t} - \frac{1 - \alpha_t}{\sqrt{1 - \bar{\alpha}_t}} \epsilon_\theta(x^{\boldsymbol{d \rightarrow f}}_{t}, t) \right) + \sigma_t \boldsymbol{z},
\end{equation}
where $\epsilon_\theta( x^{f \rightarrow d}_t, t)$  and $\epsilon_\theta(x^{d \rightarrow f}_t, t)$ are noise schemes. $\epsilon_\theta$ is the trained noise estimator, $\sigma$ is the variance, $\boldsymbol{z} \sim \mathcal{N}(0, \boldsymbol{I})$.

\textbf{Pattern-aware (PA) estimator.} To enhance the model's sensitivity to disease-semantics in both fMRI and dMRI, we incorporate a disease-informed estimator into the cross-modal denoising process. Disease-semantics are characterized by functional-structural patterns in  brain atlas regions $\mathcal{A}_f$ and $\mathcal{A}_d$. The functional patterns in fMRI can be inferred from structural patterns in dMRI, and the noise estimation, though indirect, follows similar patterns within regions $\mathcal{A}_f$ and $\mathcal{A}_d$. We align the modes between noisy fMRI/dMRI $P_x$ and their estimated noise $P_n$. 
The pattern-aware estimator operates as follows ( $\epsilon_s$ is the disease estimator):
\begin{equation}
P_x: (x)^{d \rightarrow f}_{t},  (x)^{f \rightarrow d}_{t} = \{x^{d \rightarrow f}_t  , x^{f \rightarrow d}_t\} \times \{ \mathcal{A}_f, \mathcal{A}_d\},
\end{equation}
\begin{equation}
P_n: (n)^{d \rightarrow f}_{t},  (n)^{f \rightarrow d}_{t} = \{\epsilon_s(x)^{f \rightarrow d}_{t} ,  \epsilon_s(x)^{d \rightarrow f}_{t}\} ,
\end{equation}
% \begin{align}
% L_{DA} = L_1\{P_x, P_n \mid [(x)^{d \rightarrow f}_{t},(n)^{d \rightarrow f}_{t}] + [(x)^{f \rightarrow d}_{t}, (n)^{f \rightarrow d}_{t}] \} 
% \end{align} P_x, P_n \mid 
\begin{equation}
L_{PA} =   L_1  \left[ (x)^{d \rightarrow f}_{t}, (n)^{d \rightarrow f}_{t} \right] + L_1  \left[ (x)^{f \rightarrow d}_{t}, (n)^{f \rightarrow d}_{t} \right] 
\end{equation}
% where. 
% $(x)$ denotes disease-activated region (pattern $P$) and $x$ denotes the original space.

\begin{figure}[htbp]
\vspace{-2mm}
    \centering
    \includegraphics[width=0.47\textwidth]{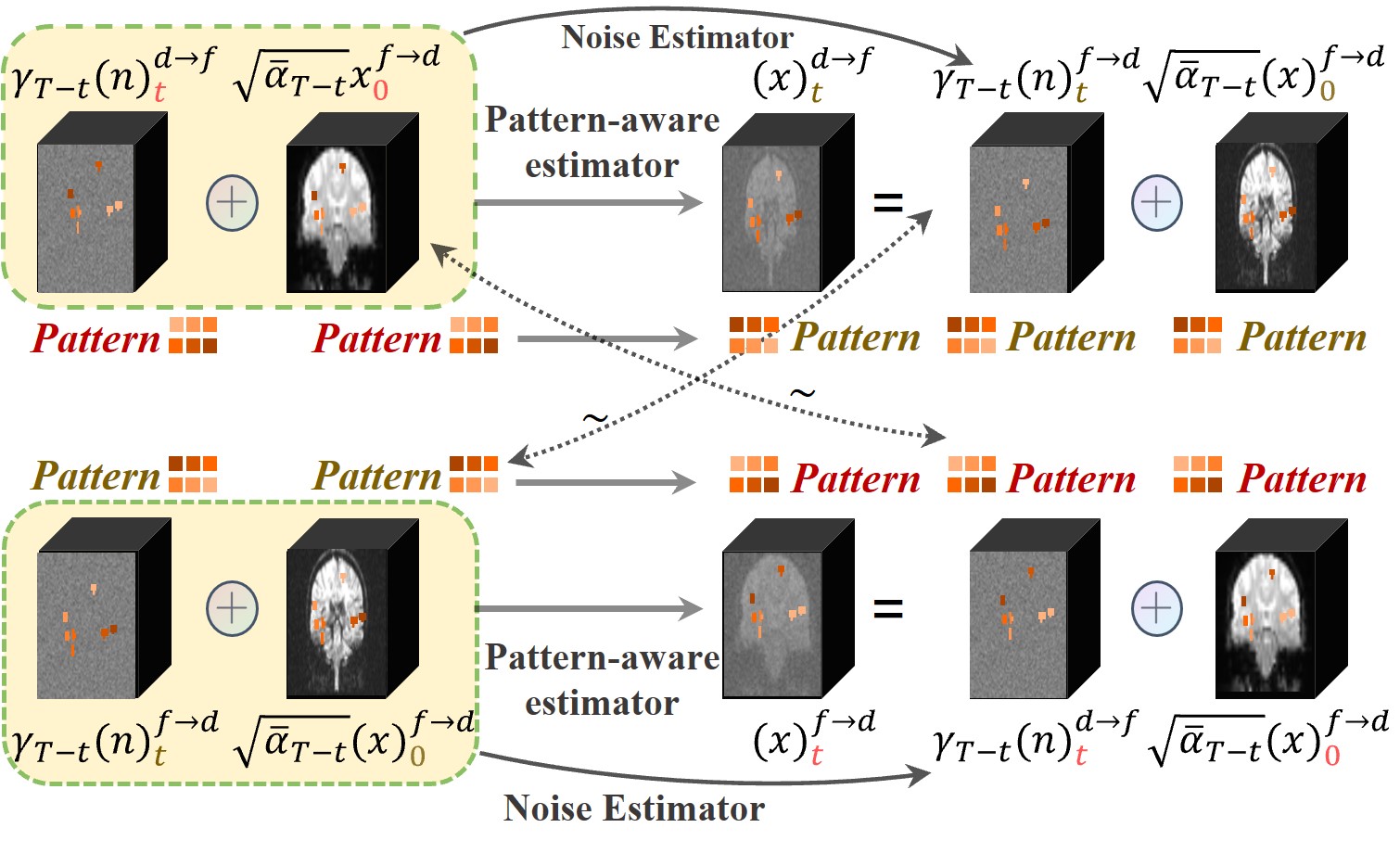} % Insert image, width set to full text width
    \vspace{-3mm}
    \caption{ The details of pattern-aware module.
    % pattern-aware Diffusion Model. Our diffusion model incorporates a noise estimator (NE) for initial fMRI and dMRI generation through progressive denoising. A pattern-aware (DA) module ensures semantic consistency  (pattern at the brain disease-related regions of atlas ) across noisy image and noise. Tissue and microstructure details are refined through a dedicated projection network and a microstructure refinement loss $\mathcal{L}_{mic}$, where $\gamma_t = \sqrt{1 - \bar{\alpha}_t}$. 
    } % Caption for the image
    \label{fig:da} % Label for referencing the image
    \vspace{-3mm}
\end{figure}
\vspace{-2mm}
\subsection{Tissue and Microstructure Refinements}

%\subsection{Tissue Refinement Network}

\textbf{Tissue Refinement (TR) Network}. To enhance spatial delineation of neuroanatomical features in global brain visualization, we implement a tissue refinement network with two distinct branches: (1) a backbone branch $\boldsymbol{B}$ and (2) a tissue projection branch $\boldsymbol{U}$. The backbone branch learns fundamental features from the fMRI and dMRI images generated by our disease-semantic-informed dual-modal diffusion model through a series of convolutional layers. The tissue projection branch utilizes a U-Net architecture to learn the tissue transformation between BOLD data space and water diffusion data space. 
%This pixel-to-pixel transformation framework enhances tissue details in the generated fMRI and dMRI by establishing direct spatial correspondences between source and target domains. 
The details of the tissue refinement network are
shown in appendix Figure 1.
The tissue refinement loss $\mathcal{L}^{d \rightarrow f}_{tis} $ and $\mathcal{L}^{f \rightarrow d}_{tis} $ can be defined as:
\begin{align}
\mathcal{L}^{d \rightarrow f}_{tis} =\ & \mathcal{L}_1\left(\boldsymbol{B}\left(x^{d \rightarrow f}_{2T}\right) + \boldsymbol{U}\left(x^{d \rightarrow f}_{0}\right), x^{f \rightarrow d}_{0}\right) \nonumber \\
& + \mathcal{L}_{mic}\left(\boldsymbol{B}\left(x^{d \rightarrow f}_{2T}\right) + \boldsymbol{U}\left(x^{d \rightarrow f}_{0}\right), x^{f \rightarrow d}_0\right),
\end{align}
% \begin{align}
% \small
% \mathcal{L}^{d \rightarrow f}_{tis} =\ & \mathcal{L}_1\left(\boldsymbol{B}\left(x^{d \rightarrow f}_{2t}\right) + \boldsymbol{U}\left(x^{f \rightarrow d}_{2t}\right), x^{d \rightarrow f}_{0}\right) \nonumber \\
% & + \mathcal{L}_m\left(\boldsymbol{B}\left(x^{d \rightarrow f}_{2t}\right) + \boldsymbol{U}\left(x^{f \rightarrow d}_{2t}\right), x^{d \rightarrow f}_0\right),
% \end{align}
\begin{align}
\mathcal{L}^{f \rightarrow d}_{tis} =\ & \mathcal{L}_1\left(\boldsymbol{B}\left(x^{f \rightarrow d}_{2T}\right) + \boldsymbol{U}\left(x^{f \rightarrow d}_{0}\right), x^{d \rightarrow f}_{0}\right) \nonumber \\
& + \mathcal{L}_{mic}\left(\boldsymbol{B}\left(x^{f \rightarrow d}_{2T}\right) + \boldsymbol{U}\left(x^{f \rightarrow d}_{0}\right), x^{d \rightarrow f}_{0}\right),
\end{align}
% where $\mathcal{L}_1$ represents L1 loss.
where $\boldsymbol{B}$ represents backbone block, $ \boldsymbol{U}$ represents U-Net architecture, $\mathcal{L}_{mic}$ represents the microstructure refinement loss and $\mathcal{L}_1$ represents L1 loss.

% \subsection{Microstructure Refinement}

\textbf{Efficient Microstructure Refinement (MR)}. 
We introduce a 3D MRI-specific microstructure refinement loss to enhance the microstructural details of fMRI and dMRI, which is inspired by the perceptual loss.
%While traditional perceptual loss projects RGB channels into high-dimensional feature spaces for image alignment, direct application to 3D volumetric data is computationally intensive and susceptible to noise. 
To achieve both effectiveness and efficiency, we project the generated and target 3D neuroimages onto 2D planes along three anatomical axes: axial, sagittal, and coronal ($Mean_c, Mean_a, Mean_c$). These multi-planar projections are then processed through a microstructure perception ($\boldsymbol{M}$) network, which maps them to a high-dimensional feature space for detail alignment between generated and target domains.  Figure 2 in the appendix illustrates our microstructure refinement loss architecture. 
The mathematical definition is as follows ($\mathcal{L}_1$ represents L1 loss):
\begin{equation}
v_g^{c}, v_g^{a}, v_g^s = \{Mean_c, Mean_a, Mean_s\} \rightarrow (x_{gen}),
\end{equation}
\begin{equation}
v_{tar}^{c}, v_{tar}^{a}, v_{tar}^s = \{Mean_c, Mean_a, Mean_s\} \rightarrow (x_{tar}),
\end{equation}
\begin{equation}
\mathcal{L}_{mic}  = \mathcal{L}_1\{\boldsymbol{M}(v_g^{c}, v_g^{a}, v_g^s),\boldsymbol{M}(v_{tar}^{c}, v_{tar}^{a}, v_{tar}^s )\},
\end{equation}
where $\boldsymbol{M}$ represents MP network (based on VGG family), $x_{gen}$ represents the generated fMRI and dMRI images, $x_{tar}$ represents target fMRI and dMRI images, $Mean_a, Mean_s, Mean_c$ represents the mean operation across axial, sagittal, coronal view.
align the microstructure detail of generated fMRI and dMRI with target fMRI and dMRI via dimensionality reduction and projection to feature space.  
 
\vspace{-3mm}
\subsection{Total Loss}
The training of PDS is divided into two stages.
: (1) the losses $L_{s1}$ of noise estimator and the pattern-informed estimator are computed and then backpropagated to generate the initial neuroimaging. 
% The calculation of the loss $L_{s1}$ can be found in the \textbf{ appendix}. 
In the second stage, the loss $L_{s2}$ of the tissue refine network is computed and then backpropagated. 
$L_{s2}$ consists of $L_1$ and $L_{mic}$ between the refined neuroimaging and the real neuroimaging. 
The model is trained in two stages: (1) generating initial neuroimaging using the dual-modal diffusion model and disease semantic refinement, and (2) tissue and microstructure refinement to improve the output from the semantic refinement stage. The first stage training loss $\mathcal{L}_{\text{s1}}$ of dual-modal diffusion model with disease semantic refinement can be defined as:
\begin{align}
\mathcal{L}^1_{s1} = \ & \mathcal{L}_1 (\epsilon_\theta(  x^{f \rightarrow d}),\mathcal{N}(0, \mathbf{I})) 
+ \mathcal{L}_1 (\epsilon_\theta( x^{d \rightarrow f}),\mathcal{N}(0, \mathbf{I})) \nonumber \\
&+ \mathcal{L}_1 (\epsilon_s(x)^{d \rightarrow f}, (x)^{d \rightarrow f}) + \mathcal{L}_1 (\epsilon_s(x)^{f \rightarrow d}, (x)^{f \rightarrow d})
\end{align}
\begin{align}
\mathcal{L}^m_{s1} = \ & \mathcal{L}_{mic} (\epsilon_\theta( x^{f \rightarrow d}),\mathcal{N}(0, \mathbf{I})) 
+ \mathcal{L}_{mic} (\epsilon_\theta( x^{d \rightarrow f}),\mathcal{N}(0, \mathbf{I})) \nonumber \\
&+ \mathcal{L}_{mic} (\epsilon_s(x)^{d \rightarrow f}, (x)^{d \rightarrow f}) + \mathcal{L}_{mic} (\epsilon_s(x)^{f \rightarrow d}, (x)^{f \rightarrow d})
\end{align}
\begin{equation}
\mathcal{L}_{s1} = \mathcal{L}^1_{s1} + \mathcal{L}^{m}_{s1} +L_{PA}
\end{equation}

The second stage train loss consists of tissue and microstructure refinement losses to improve the output fMRI and dMRI from the disease semantic refinement stage. The second stage training loss $\mathcal{L}_{\text{s2}}$ can be defined as:
\begin{equation}
\mathcal{L}_{s2} = \mathcal{L}^{f \rightarrow d}_{tis} +  \mathcal{L}^{d \rightarrow f}_{tis} + \mathcal{L}_{mic}  
\end{equation}
%\vspace{-3mm}

\begin{table*}[htbp]
\scriptsize
\centering
\caption{Demographic characteristics and MRI acquisition parameters across three cohorts: Hospital, ADNI, and OASIS. Cognitive assessments include MMSE and MoCA scores; MRI parameters include repetition time (TR), echo time (TE), field of view (FoV), slice number, flip angle, and layer thickness. Data are reported as mean $\pm$ standard deviation where available.}
\label{tab:infor}
\begin{tabular}{@{}l l r r r r r r@{}}
\toprule
\textbf{Cohort} & \textbf{Group} & \textbf{F/M} & \textbf{Weight (kg)} & \textbf{Age (years)} & \textbf{MMSE} & \textbf{MoCA} & \textbf{Education (years)} \\
\midrule

\multirow{2}{*}{Hospital} 
& NC  & 44/33 & 85.30 $\pm$ 5.34 & 64.48 $\pm$ 5.73 & 29.55 $\pm$ 0.72 & 26.50 $\pm$ 2.10 & 12.15 $\pm$ 2.94 \\
& MCI & 70/29 & 84.28 $\pm$ 5.02 & 65.31 $\pm$ 6.70 & 25.31 $\pm$ 1.04 & 21.06 $\pm$ 2.75 & 10.45 $\pm$ 2.94 \\
\addlinespace[0.5em]

& fMRI & — & — & — & — & — & TR: 2000 ms, TE: 230 ms, FoV: 100$\times$100 mm$^2$, \\
&      &   &   &   &   &   & slices: 31, FA: 90$^\circ$, thickness: 5 mm \\
& dMRI & — & — & — & — & — & TR: 6800 ms, TE: 93 ms, FoV: 256$\times$256 mm$^2$, \\
&      &   &   &   &   &   & slices: 46, FA: 90$^\circ$, thickness: 3 mm \\
\midrule

\multirow{2}{*}{ADNI} 
& NC  & 281/187 & 79.29 $\pm$ 13.09 & 76.70 $\pm$ 6.47 & 29.00 $\pm$ 1.41 & — & — \\
& MCI & 149/221 & 74.26 $\pm$ 12.39 & 73.40 $\pm$ 7.07 & 27.56 $\pm$ 1.89 & — & — \\
\addlinespace[0.5em]

& fMRI & — & — & — & — & — & TR: 2000 ms, TE: 30 ms, FoV: 93$\times$93 mm$^2$, \\
&      &   &   &   &   &   & slices: 31/48, FA: 80$^\circ$, thickness: 3.3 mm \\
& dMRI & — & — & — & — & — & TR: 6800 ms, TE: 3 ms, FoV: 93$\times$93 mm$^2$, \\
&      &   &   &   &   &   & slices: 16, FA: 9$^\circ$, thickness: 1.2 mm \\
\midrule

\multirow{3}{*}{OASIS} 
& NC  & — & — & 68.62 $\pm$ 4.73 & 28.90 $\pm$ 1.09 & — & 12.28 $\pm$ 3.53 \\
& MCI & — & — & 73.52 $\pm$ 8.39 & 27.09 $\pm$ 1.86 & — & 12.71 $\pm$ 3.13 \\
& AD  & — & — & 78.41 $\pm$ 5.39 & 23.06 $\pm$ 5.13 & — & — \\
\addlinespace[0.5em]

& fMRI & — & — & — & — & — & TR: 2000 ms, TE: 27 ms, FoV: 100$\times$100 mm$^2$, \\
&      &   &   &   &   &   & slices: 36, FA: 90$^\circ$, thickness: 4 mm \\
& dMRI & — & — & — & — & — & TR: 14500 ms, TE: 112 ms, FoV: 100$\times$100 mm$^2$, \\
&      &   &   &   &   &   & slices: 80, FA: 90$^\circ$, thickness: 2 mm \\
\bottomrule
\end{tabular}
\vspace{-3mm}
\end{table*}
\section{Experiments} 

 \subsection{Datasets}
% In this study, we used open-access datasets from the Alzheimer's Disease Neuroimaging Initiative (ADNI), the Open Access Series of Imaging Studies-3 (OASIS-3), and a private hospital-collected dataset to train the PDS. The ADNI dataset consists of 468 NC and 370 MCI patients. The OASIS-3 contains 1473 NC, 295 MCI participants, and 41 AD participants. The hospital-collected data comprises 77 NC and 99 MCI participants. To ensure sufficient training data, the fMRI and dMRI were proportionally divided along the time axis, and the average value was calculated within each time interval, resulting in 56,000 scans. For validation of generated fMRI/dMRI diagnostic performance, we utilize the synthetic and real dual-modal neuroimaging data from 173 participants for modality completion and NC/MCI/AD classification. Detailed demographics and MRI parameters for all datasets are provided in appendix Table \ref{tab:infor}.

In this study, we trained \textit{PDS} using data from ADNI, OASIS-3, and a private hospital dataset. ADNI includes 468 NC and 370 MCI participants; OASIS-3 contains 1473 NC, 295 MCI, and 41 AD; the hospital dataset comprises 77 NC and 99 MCI. To augment training, fMRI and dMRI were divided along the time axis, with averaged intervals yielding 56,000 total scans. For diagnostic validation, synthetic and real dual-modal data from 173 participants were used for modality completion and NC/MCI/AD classification. Dataset demographics and MRI parameters are summarized in Appendix.
The image quality evaluation metrics including Peak Signal-to-Noise Ratio (PSNR) and Structural Similarity Index Measure (SSIM) are employed to measure the synthesis fMRI and dMRI quality.
%The data undergo preprocessing using SPM12 \cite{Tzourio-Mazoyer2002spm}, including slice-timing correction, head motion estimation and correction, intra-subject registration, and co-registration.

 \subsection{Baselines}
% We compare our proposed method against three categories of baseline approaches:

 \textbf{GF:} This category includes GAN  \cite{kalluvila2022synthetic}, pGAN \cite{liu2019perceptual}, SA-GAN \cite{zhang2019self}, and pSAGAN \cite{liu2020attentive}. The basic GAN framework employs a U-Net architecture to construct both the generator and discriminator, utilizing the $\mathcal{L}_1$ loss to ensure the model convergence. Building upon the GAN structure, pGAN incorporates the perceptual loss into the loss function for enhancing the texture of generated images, SA-GAN introduces self-attention mechanisms into the U-Net architecture, while pSAGAN further enhances the loss function of SA-GAN by adding the perceptual loss.
  % \textbf{GF:} Includes GAN~\cite{kalluvila2022synthetic}, pGAN~\cite{liu2019perceptual}, SA-GAN~\cite{zhang2019self}, and pSAGAN~\cite{liu2020attentive}. Basic GAN uses a U-Net generator and discriminator with $\mathcal{L}_1$ loss. pGAN adds perceptual loss for texture enhancement; SA-GAN introduces self-attention to U-Net; pSAGAN combines both.

 \textbf{DF:} This category includes U-DDIM \cite{ramanarayanan2024dce}, SA-DDPM \cite{chang2024high}, LDM \cite{friedrich2025high}, and DiT \cite{hatamizadeh2024diffit} . U-DDIM uses the U-Net architecture to estimate noise at each step and employs the DDIM strategy to reduce inference time at denoising process. SA-DDPM utilizes a U-Net with self-attention to estimate noise at each step and applies the DDPM strategy for denoising. LDM estimates noise in the low-dimensional space to reduce computation cost and leverages cross-attention to improve the image quality. DiT replaces the U-Net with a transformer.
 %for noise estimation.
 % \textbf{DF:} Includes U-DDIM~\cite{ramanarayanan2024dce}, SA-DDPM~\cite{chang2024high}, LDM~\cite{friedrich2025high}, and DiT~\cite{hatamizadeh2024diffit}. U-DDIM uses DDIM for fast inference; SA-DDPM integrates self-attention into U-Net; LDM operates in low-dimensional space with cross-attention; DiT replaces U-Net with Transformer.
 
 \textbf{UF:} This approach uses a U-Net or U-Net variants to directly synthesize the target modality data \cite{kalluvila2022synthetic}. 
 % The synthesis of DS strategy is based on constructing direct mapping function.

 \textbf{Hybrid-DM \& GAN (DG).} This method combines diffusion models and GANs, where the diffusion model generates the initial images, and the GAN architecture is employed to refine the image details \cite{ozbey2023unsupervised}. 

%  \textbf{GF:} Includes GAN~\cite{kalluvila2022synthetic}, pGAN~\cite{liu2019perceptual}, SA-GAN~\cite{zhang2019self}, and pSAGAN~\cite{liu2020attentive}. Basic GAN uses a U-Net generator and discriminator with $\mathcal{L}_1$ loss. pGAN adds perceptual loss for texture enhancement; SA-GAN introduces self-attention to U-Net; pSAGAN combines both.

% \textbf{DF:} Includes U-DDIM~\cite{ramanarayanan2024dce}, SA-DDPM~\cite{chang2024high}, LDM~\cite{friedrich2025high}, and DiT~\cite{hatamizadeh2024diffit}. U-DDIM uses DDIM for fast inference; SA-DDPM integrates self-attention into U-Net; LDM operates in low-dimensional space with cross-attention; DiT replaces U-Net with Transformer.

% \textbf{UF:} Direct modality synthesis using U-Net or variants~\cite{kalluvila2022synthetic}.

% \textbf{Hybrid DM \& GAN (DG):} Combines diffusion models and GANs, where the former generates initial images and the latter refines details~\cite{ozbey2023unsupervised}.
\vspace{-2mm}
 \subsection{Experiment Setting}
 % \section{Experiment Setting}
\label{sec: setting}
\textbf{Implementation Details.} The framework is implemented in PyTorch~2.4 with CUDA~12.1 and cuDNN~8.9, trained on 8 NVIDIA L20 GPUs (48\,GB VRAM each) using distributed data parallelism (batch size = 1 per GPU). The disease-semantics-enhanced dual-modal 3D diffusion model employs a 3D self-attention U-Net with four hierarchical blocks (channel dimensions [64, 128, 256, 512]), each containing two residual units of four $3\times3\times3$ convolutions, and four parallel attention heads~\cite{vaswani2017attention} for cross-modal fusion. Trained over $T=1000$ timesteps ($\gamma=0.996$) in two stages—300 epochs for baseline and semantics learning, followed by 150 epochs for refinement—it requires 420 hours for diffusion training and 84 hours for tissue/microstructure refinement using AdamW ($\text{lr}=10^{-5}$, weight decay=$5\times10^{-6}$) with mixed-precision (FP16) and gradient scaling. The tissue refinement module uses a dual-branch architecture: a backbone with two $3\times3\times3$ conv layers and a 3D U-Net projection branch (same block structure as the diffusion model) to align fMRI and dMRI in a shared embedding space via MSE loss. The microstructure refinement module aggregates features via mean pooling across x/y/z axes, concatenates the three views, and processes them through a VGG16-based perception network, which achieved optimal synthesis performance among VGG variants.
% \vspace{-2mm}
\subsection{MRI Parameter and Demographic Information}

\begin{figure*}[ht]
    \centering
    \includegraphics[width=0.99\textwidth]{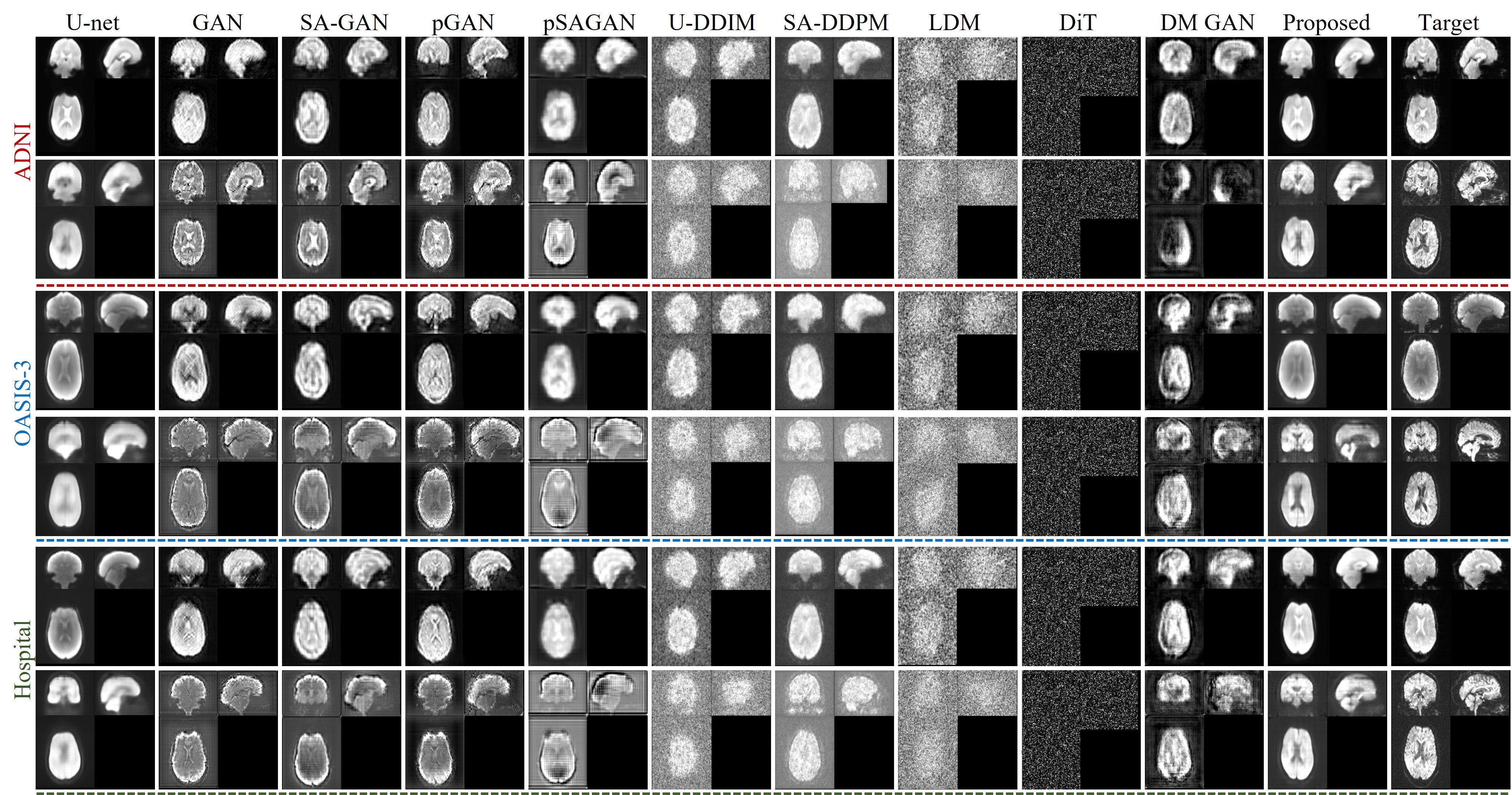} % Insert image, width set to full text width
    \vspace{-1mm}
    \caption{%Visual comparison results (axial, sagittal, and coronal view) between baseline methods and the proposed method are shown. The synthesized fMRI and dMRI images generated by the proposed method exhibit superior tissue and microstructure details. Additionally, the generated images from the proposed method are highly similar to the target images.
    Qualitative comparison across axial, sagittal, and coronal projections demonstrates our method's superior synthesis capabilities, achieving enhanced anatomical precision and improved tissue microstructure alignment with ground-truth.} % Caption for the image
    \label{fig:4} % Label for referencing the image
    \vspace{-2mm}
\end{figure*}

\begin{figure*}[ht]
    \centering
    \includegraphics[width=0.99\textwidth]{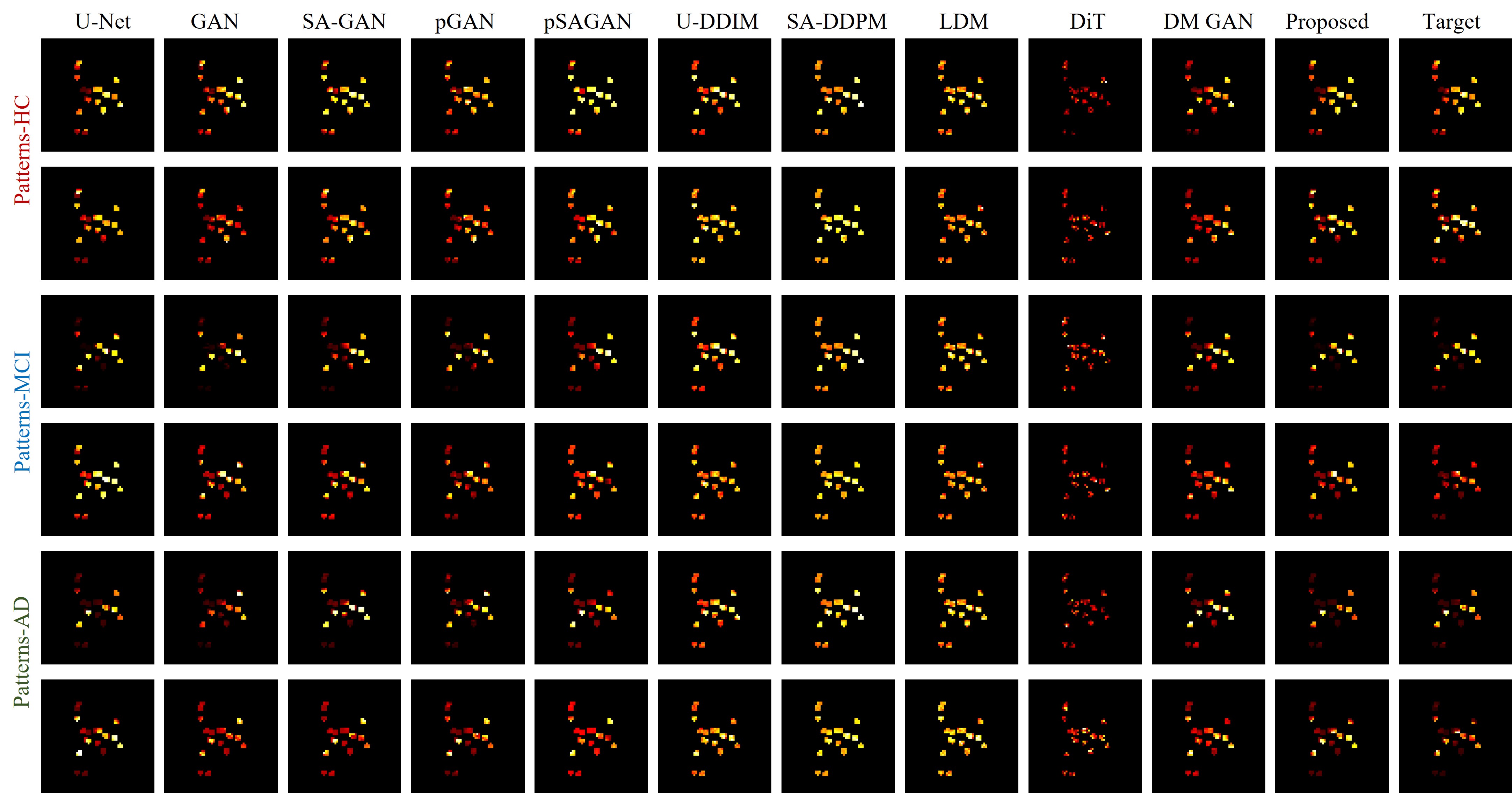} % Insert image, width set to full text width
        \vspace{-1mm}
    \caption{%Visual comparison results (axial, sagittal, and coronal view) between baseline methods and the proposed method are shown. The synthesized fMRI and dMRI images generated by the proposed method exhibit superior tissue and microstructure details. Additionally, the generated images from the proposed method are highly similar to the target images.
      Visualization comparison pattern results of roi signals (fMRI/dRMI) in different kinds of cognitive status} % Caption for the image
    \label{fig:visilization roi signals} % Label for referencing the image
    \vspace{-3mm}
\end{figure*}

The detailed information of MRI parameters and demographic information is shown in Table. \ref{tab:infor}. Both the training and validation processes are conducted on clinical datasets to address the clinical modality absence problem. This Table. \ref{tab:infor} presents the demographic information and MRI parameters of three datasets: the hospital dataset, ADNI, and OASIS. In terms of demographic information, it includes the gender ratio (F/M indicating the number of females/males), weight, age, Mini - Mental State Examination scores (MMSE), Montreal Cognitive Assessment scores (MoCA), and years of education for different neurodegenerative stage groups (NC, MCI, and AD).
For the MRI parameters, information such as repetition time, echo time, percentage of field of view, number of slices, flip angle, and slice thickness is listed separately for fMRI and dMRI. 
These data are helpful for studying the associations between the cognitive decline stage of different populations and MRI characteristics.

%\vspace{-4mm}
\section{Results}
\subsection{Quantitative and Visualization Results}
\begin{table}[h]
\centering
\small
\caption{
% Comparative analysis across three datasets demonstrates the proposed method's superiority performance in both PSNR (+9.0/4.13 dB) and SSIM (+33.31/14.96\%) over baseline methods for dual neuroimaging modalities.
Comparative analysis across three datasets demonstrates the proposed method's superiority performance. 
%in both PSNR (+1.54/1.02 dB) and SSIM (+4.12/2.2\%) over baseline methods for dual neuroimaging modalities.
}
\begin{tabular}{cccccc}%cccccc
\hline
{}& {}& \multicolumn{2}{c}{fMRI}& \multicolumn{2}{c}{dMRI}\\ \cline{3-6}
{}& {}& PSNR& SSIM& PSNR& SSIM\\ \hline
\multicolumn{1}{c}{\begin{tabular}[c]{@{}c@{}}UF\end{tabular}}& \multicolumn{1}{c}{U-Net 
 \cite{kalluvila2022synthetic} }  & 28.29& 86.72& 28.98& 75.35\\ \hline
\multicolumn{1}{c}{}& \multicolumn{1}{c}{GAN \cite{kalluvila2022synthetic}}& 20.83& 57.53& 25.38& 58.70\\
\multicolumn{1}{c}{}& \multicolumn{1}{c}{pGAN \cite{liu2019perceptual}} & 20.15& 49.27& 25.04& 56.12\\
\multicolumn{1}{c}{}& \multicolumn{1}{c}{SAGAN \cite{zhang2019self}} & 19.04& 44.59& 25.87& 62.59\\
\multicolumn{1}{c}{\multirow{-4}{*}{\begin{tabular}[c]{@{}c@{}}GF\end{tabular}}}& \multicolumn{1}{c}{pSAGAN \cite{liu2020attentive}}& 18.42& 34.93& 25.35& 58.94\\ \hline
% \multicolumn{1}{c|}{\begin{tabular}[c]{@{}c@{}}DS\end{tabular}}& \multicolumn{1}{c|}{U-Net 
%  \cite{kalluvila2022synthetic} }  & 28.29& 86.72& 29.98& 76.35\\ \hline

\multicolumn{1}{c}{}& \multicolumn{1}{c}{U-DDIM \cite{ramanarayanan2024dce}} & 10.54 & 10.67 & 10.94 & 4.67 \\
\multicolumn{1}{c}{}& \multicolumn{1}{c}{SA-DDPM \cite{chang2024high}} & 11.68 & 16.31& 7.34& 8.33\\
\multicolumn{1}{c}{}& \multicolumn{1}{c}{LDM \cite{friedrich2025high}} & 9.06& 3.65& 7.98& 1.42\\
\multicolumn{1}{c}{\multirow{-4}{*}{\begin{tabular}[c]{@{}c@{}}DF\end{tabular}}} & \multicolumn{1}{c}{DiT \cite{hatamizadeh2024diffit}}& 7.12& 1.71& 6.95& 1.99\\ \hline
\multicolumn{1}{c}{\begin{tabular}[c]{@{}c@{}}DG\end{tabular}}& \multicolumn{1}{c}{ADM \cite{ozbey2023unsupervised}   }                    & 18.66& 33.25& 25.37& 54.09\\ \hline
 \multicolumn{1}{c}{ }&  \multicolumn{1}{c}{PDS}&    \textbf{29.83}&   \textbf{90.84}&   \textbf{30.00}&   \textbf{77.55}\\ \hline
\end{tabular}
\label{tab:1}
\vspace{-3mm}
\end{table}

\begin{figure*}[ht]
    \centering
    \includegraphics[width=0.99\textwidth]
    {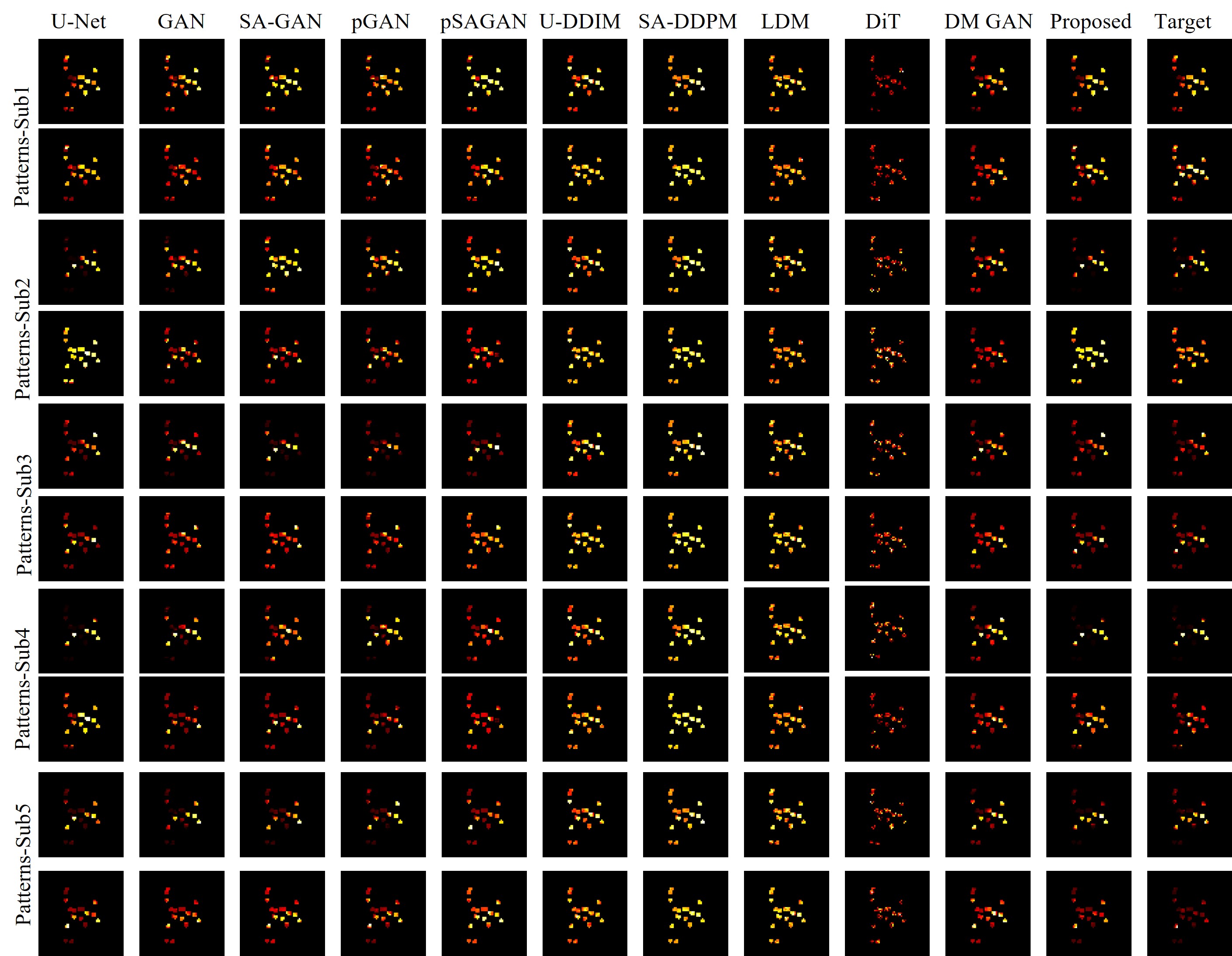} % Insert image, width set to full text width
    \vspace{-1mm}
    \caption{%Visual comparison results (axial, sagittal, and coronal view) between baseline methods and the proposed method are shown. The synthesized fMRI and dMRI images generated by the proposed method exhibit superior tissue and microstructure details. Additionally, the generated images from the proposed method are highly similar to the target images.
      Visualization comparison pattern results of roi signals (fMRI/dRMI) in subjects} % Caption for the image
    \label{fig: visilization subs roi} % Label for referencing the image
    \vspace{-5mm}
\end{figure*}

\begin{table*}[ht]

\setlength{\tabcolsep}{2mm}
\small
\centering
\caption{
% Cross-dataset evaluations demonstrate our method's SOTA performance, achieving PSNR(dB)/SSIM(\%) scores of (+9.06/33.88, +8.92/32.67) for fMRI and (+4.25/14.41, +4.48/15.81) for dMRI on OASIS-3/ADNI datasets, with hospital data reaching +8.86/25.71 (PSNR[dB]/SSIM[\%]) for fMRI and +6.5/8.66 for dMRI.
Cross-dataset evaluations demonstrate our method's SOTA performance,
achieving PSNR(dB)/SSIM(\%) scores of (+1.95/5.19, +1.68/5.17) for fMRI and (+1.21/1.37, +1.52/2.02) for dMRI on OASIS-3/ADNI datasets, with hospital data reaching +1.03/2.61 (PSNR[dB]/SSIM[\%]) for fMRI and +4.0/9.46 for dMRI.
}
\vspace{-1mm}
\begin{tabular}{cccccccccccccc}
\hline
&& \multicolumn{4}{c}{OASIS-3}& \multicolumn{4}{c}{ADNI}& \multicolumn{4}{c}{Hospital}\\ \cline{3-14} 
&& \multicolumn{2}{c}{fMRI}& \multicolumn{2}{c}{dMRI}& \multicolumn{2}{c}{fMRI}& \multicolumn{2}{c}{dMRI}& \multicolumn{2}{c}{fMRI}& \multicolumn{2}{c}{dMRI}\\ \hline
\multicolumn{1}{c}{}& \multicolumn{1}{c}{}& PSNR& SSIM& PSNR& SSIM& PSNR& SSIM& PSNR& SSIM& PSNR& SSIM& PSNR& SSIM\\ \hline
\multicolumn{1}{c}{UF}& \multicolumn{1}{c}{U-Net  \cite{kalluvila2022synthetic} }& 28.01& 86.19& 29.09& 75.72& 27.98& 85.06& 28.64& 76.28& 29.48& 89.74& 28.37& 68.09\\ \hline
\multicolumn{1}{c}{}& \multicolumn{1}{c}{GAN  \cite{kalluvila2022synthetic} }& 20.90& 57.50& 25.52& 58.89& 20.74& 57.56& 25.23&58.49& 21.65& 66.64& 25.52& 67.35\\
\multicolumn{1}{c}{}& \multicolumn{1}{c}{pGAN  \cite{liu2019perceptual} }& 20.08& 48.45& 25.11& 56.21& 20.22& 50.17& 24.96& 56.02& 21.28& 55.71& 25.22& 59.89\\
\multicolumn{1}{c}{}& \multicolumn{1}{c}{SAGAN   \cite{zhang2019self} }& 19.01& 43.22& 26.05& 62.68& 19.08& 46.07& 25.68& 62.49& 20.50& 64.60& 25.87& 68.89\\
\multicolumn{1}{c}{\multirow{-4}{*}{GF}}& \multicolumn{1}{c}{pSAGAN  \cite{liu2020attentive} }   & 18.41& 33.60& 25.58& 59.20& 18.40& 35.90& 25.09& 58.50& 19.25& 53.67& 25.25& 65.19\\ \cline{1-14} 
\multicolumn{1}{c}{{  }}& \multicolumn{1}{c}{U-DDIM  \cite{ramanarayanan2024dce}}& 10.54& 10.74& 10.94& 4.67& 10.54& 10.59& 10.94& 4.65& 10.86& 10.60& 11.40& 5.65\\
\multicolumn{1}{c}{{  }}& \multicolumn{1}{c}{SA-DDPM \cite{chang2024high} }   & 12.33& 23.09& 8.20& 9.71& 12.28& 23.92& 8.22& 9.82& 13.18& 35.66& 8.70& 14.07\\
\multicolumn{1}{c}{{  }}& \multicolumn{1}{c}{LDM  \cite{friedrich2025high} }& 9.04& 3.63& 7.98& 1.42& 9.08& 3.68& 7.97& 1.42& 9.06& 3.43& 8.26& 1.71\\
\multicolumn{1}{c}{\multirow{-4}{*}{{  DF}}} & \multicolumn{1}{c}{DiT \cite{hatamizadeh2024diffit} }& 7.06& 1.65& 6.99& 2.01& 7.18& 1.76& 6.91& 1.97& 7.36& 1.92& 6.89& 2.60\\ \hline
% \multicolumn{1}{c|}{UF}& \multicolumn{1}{c|}{U-Net  \cite{kalluvila2022synthetic} }& 28.01& 86.19& 30.09& 76.72& 27.98& 85.06& 29.64& 77.28& 29.48& 89.74& 28.37& 68.09\\ \hline
\multicolumn{1}{c}{DG}& \multicolumn{1}{c}{ADM  \cite{ozbey2023unsupervised}  }   & 18.66& 33.25& 25.37& 54.09& 18.66& 33.25& 25.37& 54.09& 18.66& 33.25& 25.37& 54.09\\ \hline
\multicolumn{1}{c}{  }& \multicolumn{1}{c}{  PDS} &   {\textbf{29.96}} &  {\textbf{91.38}} &   {\textbf{30.30}} &   {\textbf{77.09}} &   {\textbf{29.66}} &   {\textbf{90.23}} &   {\textbf{30.16}} &   {\textbf{78.30}} &   {\textbf{30.51}} &   {\textbf{92.35}} &{\textbf{32.37}} &  {\textbf{77.55}} \\ \hline
\end{tabular}

\label{tab:2}
\end{table*}

\begin{figure*}[htbp]
    \centering
    \includegraphics[width=\textwidth, height=0.7\textheight, keepaspectratio]{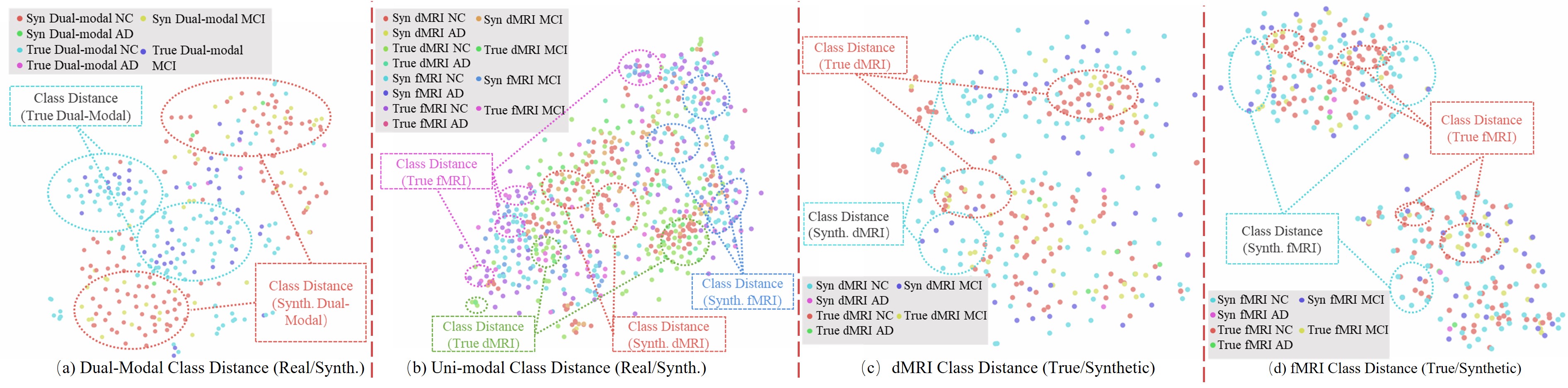} % Insert image, width set to full text width
\vspace{-3mm}
    \caption{
    t-SNE visualization of dual-modal and uni-modal data demonstrates increased inter-class separation (NC, MCI, and AD) in synthesized images compared to real images, resulting in improved classification performance as observed in Table \ref{tab:diagnosis}. 
    % The same conclusion can be inferred from appendix Fig. \ref{fig:S2}. Dual-modal t-SNE representations are constructed through principal component analysis of combined fMRI and dMRI features.
    } % Caption for the image
        \vspace{-3mm}
    \label{fig:5} % Label for referencing the image
\end{figure*}

\begin{figure}[ht]
\vspace{-3mm}
    \centering
    \includegraphics[width=0.5\textwidth]{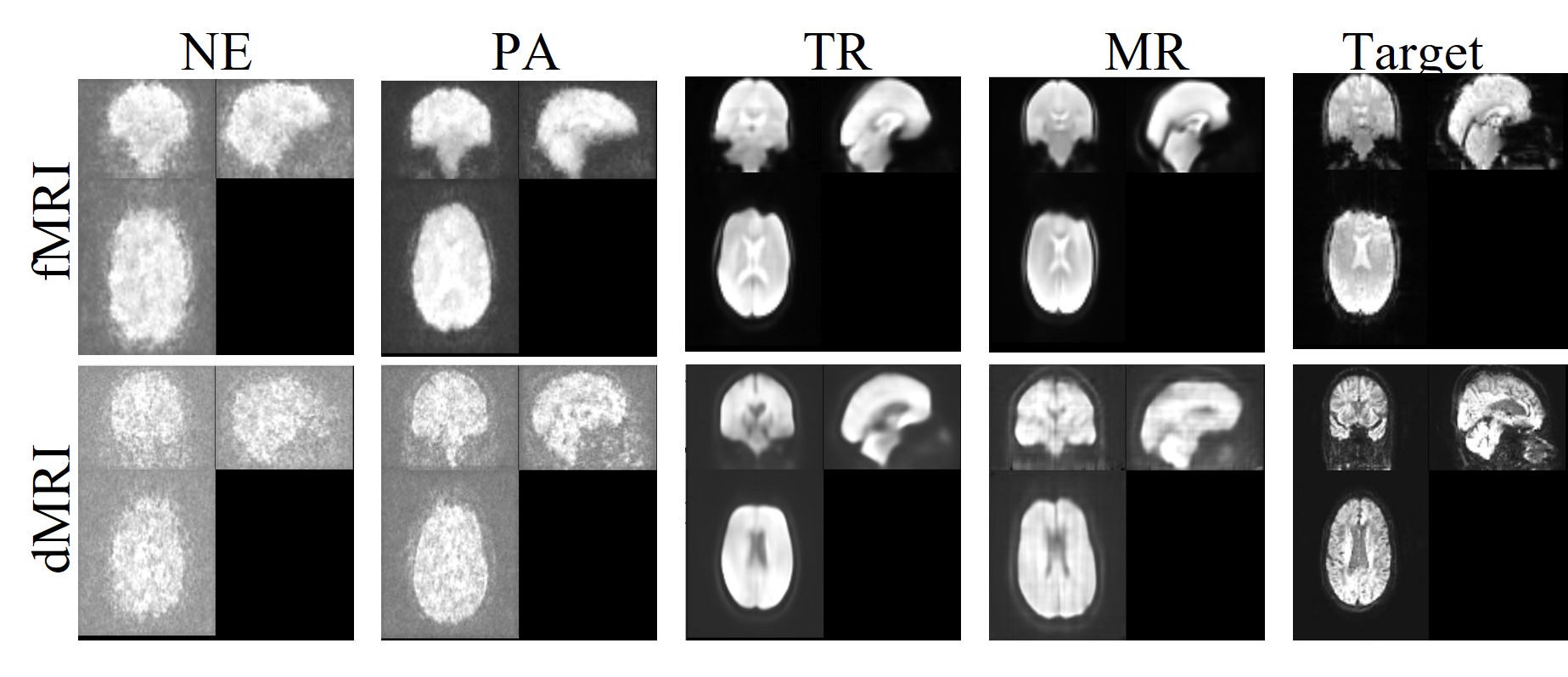} % Insert image, width set to full text width
    \vspace{-7mm}
    \caption{Visualize ablation study results of PDS.
    % disease semantic-informed module, tissue and microstructure refinement. After introducing disease semantic-informed module, TR, and MR, the tissue and microstructure details are increasing progressively.
    } % Caption for the image
    %  {-1mm}
    \label{fig:6} % Label for referencing the image
    \vspace{-1mm}
\end{figure}

\begin{table}[htbp]
\centering
\footnotesize
\vspace{-1mm}
\caption{Ablation study of disease-semantic-informed, tissue and microstructure refinements. 
% With  disease-semantic-informed module, tissue refinement (TR), and microstructure refinement (MR), the PSNR (dB) and SSIM(\%) are increased progressively.
}
\vspace{-1mm}
\begin{tabular}{cccccccc}
\hline
\multicolumn{2}{c}{\begin{tabular}[c]{@{}c@{}}Dual-modal  \\ 3D diffuser\end{tabular}}& \multicolumn{2}{c}{\begin{tabular}[c]{@{}c@{}}Refine-  \\ ment\end{tabular}}& \multicolumn{2}{c}{fMRI} & \multicolumn{2}{c}{dMRI} \\ \hline
\multicolumn{1}{c}{\begin{tabular}[c]{@{}c@{}}NE \end{tabular}} & \multicolumn{1}{c}{\begin{tabular}[c]{@{}c@{}}PA\end{tabular}} & \begin{tabular}[c]{@{}c@{}}TR\end{tabular} & \begin{tabular}[c]{@{}c@{}}MR\end{tabular} & PSNR & SSIM& PSNR& SSIM\\ \hline
   $\checkmark$ &  $\times $ & $\times $ & $ \times$ & 11.68 & 16.31& 7.34& 8.33 \\
     $\checkmark$  &    $\checkmark$ &  $\times $  &   $\times $  & 12.21 & 22.44 & 8.46 & 10.48 \\
     $\checkmark$ &    $\checkmark$ &    $\checkmark$  &  $\times $& 28.47 & 86.56 & 27.98& 76.42 \\
   $\checkmark$ &    $\checkmark$ &    $\checkmark$  &     $\checkmark$&    29.83&    90.84&    30.00&    77.55\\ \hline
\end{tabular}

\label{tab:4}
\vspace{-2mm}
\end{table}

%  {-3mm}
%The metrics, including peak signal-to-noise ratio (PSNR) and structural similarity index measure (SSIM), are employed to evaluate the synthesized fMRI and dMRI image quality. The total mean validation results across three datasets are presented in Table \ref{tab:1}. The validation results for each dataset are detailed in Table \ref{tab:2}. The visualization results are illustrated in Figure \ref{fig:4}. Both the quantitative results and visualizations demonstrate that the indirect estimation strategy in baseline category DMF exhibits inferior synthesis performance compared to the other synthesis strategy. Similarly, the generator and discriminator architectures in the GF category fail to achieve satisfied performance, primarily due to the model collapse issue. Likewise, the refinement method ADM in the DMG category encounters the same problem as GF, resulting in suboptimal performance. 
Synthesis quality of fMRI/dMRI images was evaluated using peak signal-to-noise ratio (PSNR) and structural similarity index measure (SSIM). Cross-dataset averaged results (Table \ref{tab:1}) and per-dataset validations (Table \ref{tab:2}) reveal three critical shortcomings: (1) DF's indirect estimation strategy yields suboptimal performance (Fig. \ref{fig:4} visualizations); (2) GF's GAN architectures exhibit model collapse; (3) ADM inherits GF's failure patterns. Both quantitative metrics and visual inspection confirm these limitations.
%The U-Net method in the baseline category DS achieves better performance than the DF, GF, and DMG categories. However, the U-Net demonstrates weaker generalization ability compared to the proposed method when processing large-scale datasets (specifically, 51,000 scans in this study), leading to inferior performance against the proposed method. 
%Compared with exxisting generative architecture, proposed method can achieve 9dB (PSNR) and 33.31 \% (SSIM) for fMRI synthesis and  4.13dB (PSNR) and 14.96 \% (SSIM) for dMRI synthesis in the dataset contains all three dataset. While for individual datasset, 8.68dB (PSNR) and 33.38 \% (SSIM) for fMRI synthesis and  4.25dB (PSNR) and 14.41 \% (SSIM) for dMRI synthesis on OASIS-3, 8.92dB (PSNR) and 32.67 \% (SSIM) for fMRI synthesis and  4.48dB (PSNR) and 15.81 \% (SSIM) for dMRI synthesis on ADNI, 8.86dB (PSNR) and 25.71 \% (SSIM) for fMRI synthesis and  6.5dB (PSNR) and 10.2 \% (SSIM) for dMRI synthesis on Hospital.
% Our PDS demonstrates substantial improvements over existing architectures. On the combined dataset, we achieve gains of 9.00 dB in PSNR and 33.31\% in SSIM for fMRI synthesis, alongside improvements of 4.13 dB and 14.96\% for dMRI synthesis. These performance gains are consistent across individual datasets, with our framework consistently outperforming baselines in both structural fidelity (PSNR) and perceptual quality (SSIM). Hence, PDS establishes SOTA performance in  quantitative metrics and anatomical details.
Our PDS demonstrates substantial improvements over existing architectures. On the combined dataset, we achieve gains of 1.54 dB in PSNR and 4.12\% in SSIM for fMRI synthesis, alongside improvements of 1.02 dB and 2.2\% for dMRI synthesis. These performance gains are consistent across individual datasets, with our framework consistently outperforming baselines in both structural fidelity (PSNR) and perceptual quality (SSIM). Hence, PDS establishes SOTA performance in  quantitative metrics and anatomical details.

\begin{table*}[h]
\footnotesize

\centering
\caption{
Diagnostic performance for normal controls (NC) / mild cognitive impairment (MCI) /Alzheimer's disease (AD) was assessed across all scenarios using synthesized fMRI and dMRI data. Models show bidirectional prediction accuracy: synthetic-trained models effectively predict real data (and vice versa), with enhanced cross-scenario consistency. 
For example, Scheme S5 (synthetic test data) vs S8 (real test data) comparisons reveal +13.21\% ACC/+6.07\% AUC/+6.33\% F1-score. Cross-testing (S1 [synthetic fMRI test] vs S4 [synthetic dMRI test] ) demonstrates fMRI's superior diagnostic accuracy (+9.45\% ACC), driven by its +13.29\% SSIM advantage over synthetic dMRI.
}
\vspace{-1mm}

\begin{tabular}{cccccccccccccc}
\hline
& \multicolumn{4}{c}{Train}    & \multicolumn{4}{c}{ Test}   & \multicolumn{5}{c}{ Metrics}\\ \hline
& \multicolumn{2}{c}{ fMRI} & \multicolumn{2}{c}{ dMRI} & \multicolumn{2}{c}{ fMRI} & \multicolumn{2}{c}{ dMRI} &   &  &   \\ \cline{2-9}
& Syn& True & Syn& True & Syn& True & Syn& True & \multirow{-2}{*}{ ACC} & \multirow{-2}{*}{ PRE} & \multirow{-2}{*}{ REC} & \multirow{-2}{*}{ F1-score} &  \multirow{-2}{*}{ AUC} \\ \hline
Scheme 1  & $\checkmark$ &$\times$ & $\times$  & $\checkmark$ & $\times$ &$\checkmark$  & $\times$  & $\checkmark$ &56.60& 25.67& 28.34 & 26.64&45.22  \\
   Scheme S1  &  $\checkmark$ &  $\times$ &  $\times$  &  $\checkmark$   &  $\times$ &  $\checkmark$&   $\checkmark$ &  $\times$ &  54.70&   30.08&   31.41 &   31.12&  48.21 \\
 Scheme S2 & $\checkmark$ &$\times$ & $\times$  & $\checkmark$ & $\checkmark$ &  $\times$& $\checkmark$  & $\times$ &56.60& 28.28& 29.66 & 33.84&46.51  \\
  Scheme S3 & $\checkmark$ &$\times$ & $\times$  & $\checkmark$ & $\checkmark$ &$\times$ & $\times$  & $\checkmark$  &64.15& 34.59& 34.59 & 33.38 &51.26  \\
Scheme 2 & $\times$& $\checkmark$ & $\checkmark$ & $\times$& $\times$ & $\checkmark$  & $\times$ &  $\checkmark$  &67.92 & 40.99 & 41.68 & 41.33&58.14 \\
   Scheme S4&   $\times$&   $\checkmark$ &  $\checkmark$ &  $\times$&   $\checkmark$&   $\times$ &   $\times$ &   $\checkmark$ &  64.15 &   29.58 &   31.95 &   29.21&  48.68 \\
   Scheme S5&   $\times$&   $\checkmark$ &   $\checkmark$ &   $\times$&    $\checkmark$&    $\times$ &   $\checkmark$  &  
$\times$&  67.92&   39.05&   37.71 &   37.44&  54.28  \\
 Scheme S6 & $\times$& $\checkmark$ & $\checkmark$ & $\times$& $\times$& $\checkmark$ & $\checkmark$ & $\times$  &77.35 & 48.07 & 48.82 & 48.44&65.65 \\
Scheme 3  & $\checkmark$ & $\times$ &  $\checkmark$& $\times$  & $\times$ &  $\checkmark$  &    $\times$ &  $\checkmark$  & 58.49 & 33.84& 34.53&34.17 & 51.23   \\
 Scheme S7& $\checkmark$ & $\times$ &  $\checkmark$& $\times$ & $\checkmark$ &  $\times$ & $\times$ &  $\checkmark$  &58.49 & 32.60 &33.21 &32.87 &49.95 \\
   Scheme S8 &   $\checkmark$ &   $\times$ &    $\checkmark$&   $\times$ &   $\times$ &   $\checkmark$  &    $\checkmark$&    $\times$  &  54.71 &   30.88 &   31.41 &   31.11&  48.21 \\
 Scheme S9&  $\checkmark$ & $\times$ &  $\checkmark$& $\times$&  $\checkmark$ & $\times$ &  $\checkmark$& $\times$  &56.60 & 30.19 & 30.91 & 30.51&47.79 \\
Scheme 4 & $\times$& $\checkmark$ & $\times$ & $\checkmark$ & $\checkmark$ & $\times$ & $\times$ & $\checkmark$ & 64.15 & 34.59 & 34.59 & 33.84&51.62  \\
Scheme 5 &  $\times$ & $\checkmark$ & $\times$ & $\checkmark$ & $\times$ & $\checkmark$ & $\checkmark$ & $\times$ & 66.03 & 36.20 & 35.49 & 34.71&52.12 \\
Scheme 6 & $\times$ & $\checkmark$ & $\times$ & $\checkmark$ & $\checkmark$ & $\times$&  $\checkmark$   & $\times$ & 62.26 & 33.25 & 33.69 & 33.00&50.39 \\
 Scheme S10 & $\times$ & $\checkmark$ & $\times$ & $\checkmark$ & $\times$ & $\checkmark$ & $\times$ & $\checkmark$ & 69.81 & 41.47 & 39.93 & 40.00&56.43 \\ \hline
\end{tabular}
\label{tab:diagnosis}

\vspace{-1mm}
\end{table*}

% \usepackage{multirow}
% \usepackage{multirow}
% \usepackage{multirow}

\begin{table}[ht]
\setlength{\tabcolsep}{1mm}
\footnotesize
\centering
\vspace{-2mm}
\caption{Our analysis shows MR loss boosts efficiency versus standard 3D perceptual loss frameworks.}
\vspace{-1mm}
\begin{tabular}{cccc}

\hline
& 
\makecell[c]{fMRI \\ $(64 \times 64 \times 36)$ } & 
\makecell[c]{dMRI \\ $(96 \times 96 \times 60)$} & 
Total \\ \hline
\multicolumn{4}{l}{\textbf{Parameters}} \\
\makecell[c]{Conventional 3D\\ Perceptual Loss} & 
132.31M & 291.60M & 423.91M \\ 

MR loss  &   
\textbf{98.64M} &   
\textbf{140.51M} &   
\textbf{239.05M} \\ \hline
\multicolumn{4}{l}{\textbf{Inference Time}} \\
pGAN  &   
3.447s &   
3.447s &   
6.894s \\ 
PDS &   
\textbf{2.090s} &   
\textbf{2.090s} &   
\textbf{4.180s} \\ \hline
\end{tabular}
\label{tab:parameter}
\end{table}

% \vspace{-2mm}

\subsection{Synthetic Image Diagnosis}

To validate clinical utility, we conducted a three-class (NC/MCI/AD) diagnostic task using synthetic and real neuroimaging data. We implemented bidirectional validation: Schemes 1–3 train on synthetic and test on real data; Schemes 4–6 reverse this (Table~\ref{tab:diagnosis}). Both show robust performance, with improved class separation visualized via t-SNE in dual- and uni-modal spaces (Fig.~\ref{fig:5}), driven by the superior discriminability of synthetic features. Key findings include:
\begin{itemize}
    \item \textbf{Enhanced class separability}: Synthesized data shows larger inter-class distances in t-SNE visualizations.
    \item \textbf{Superior diagnostic generalization}:
    \begin{itemize}
        \item Test on synthetic: Scheme 4/6/S5 achieved ACC $64.15\%$ (fMRI), $62.26\%$ (dMRI), $67.92\%$ (both).
        \item Train on synthetic: Scheme 1/3/S8–S9 achieved ACC $56.60\%$ (fMRI), $58.49\%$ (dMRI), $54.71\%$–$56.60\%$ (both).
    \end{itemize}
\end{itemize}
Notably, higher-quality fMRI synthesis (+4.12\% SSIM) outperforms dMRI in diagnostic accuracy under synthetic testing (Scheme S4: +0.54\% F1, +9.45\% ACC vs. Scheme S1;).

% \usepackage{multirow}

\subsection{Ablation Study}
%  {-1mm}

%  {-4mm}

% We conducted comprehensive ablation studies examining the contributions of noise estimator (NE) and pattern-aware(DA) in the diffusion model, tissue refinement, and microstructure refinement. 
% The quantitative and visualization results of the ablation study are presented in Table \ref{tab:4} and Fig. \ref{fig:6}, respectively. 
% Ablation results demonstrate progressive improvements in both quantitative metrics (PSNR, SSIM) and qualitative characteristics (tissue and microstructure details) with the introduction of each component (Table \ref{tab:4}, Fig. \ref{fig:6}). Importantly, while the disease-semantic-informed module shows modest PSNR/SSIM improvements, it delivers substantial diagnostic gains (+6.95\% ACC, +6.98\% REC; appendix Table \ref{tab:S3}), underscoring its clinical significance.

% We conducted ablation studies to evaluate the contributions of the noise estimator (NE) and pattern-aware (DA) components in diffusion modeling, tissue refinement, and microstructure enhancement. Quantitative and visual results are shown in Table~\ref{tab:4} and Fig.~\ref{fig:6}, respectively. Results show progressive improvements in both PSNR/SSIM metrics and visual details with each added component. Notably, while DI yields modest quantitative gains, it significantly improves diagnostic performance (+6.95\% ACC, +6.98\% REC;), highlighting its clinical value.

Although the limitation improvement in PSNR and SSIM when introducing disease semantic-informed module, the diagnostic performance improvement (+3.77\% ACC and +6.98\% REC in table. \ref{tab:S3}) demonstrated the importance of disease semantic-informed module. 
n ablation study investigating the effects of noise estimator (NE) and pattern-aware(PA) in the PDM, tissue refinement, and microstructure refinement was conducted on the datasets. As shown in Table \ref{tab:4} and Fig. \ref{fig:6}, all the introducing module elevates PSNR/SSIM while enhancing anatomical details.
%Notably, the disease semantic-informed module drives diagnostic performance gains (+3.77\% ACC/+6.98\% REC in Table \ref{tab:S3}) despite modest metric increases, validating its clinical value.
\begin{table}[htbp]
\footnotesize
\centering
\vspace{-1mm}
\caption{The PDM outperforms conventional DFs
. 
%achieving +0.53 dB PSNR/+6.13\% SSIM for fMRI and +2.15\% SSIM for dMRI synthesis versus baselines. 
%Significant SSIM gains (S1 vs. S4) indicate enhanced diagnostic potential.
% 
}
\vspace{-1mm}
\begin{tabular}{cccccc}

\hline
    & & \multicolumn{2}{c}{fMRI} & \multicolumn{2}{c}{dMRI} \\ \cline{3-6} 
 & & PSNR     & SSIM    & PSNR      & SSIM   \\ \hline
\multicolumn{1}{c}{\multirow{4}{*}{DF}} & U-DDIM & 10.54 & 10.67& 10.94 & 4.67 \\
\multicolumn{1}{c}{}& SA-DDPM& 11.68& 16.31& 7.34& 8.33\\
\multicolumn{1}{c}{}& LDM& 9.06& 3.65& 7.97& 1.42\\
\multicolumn{1}{c}{}& DiT& 7.12& 1.71& 6.95& 1.99\\ \hline
   \multicolumn{1}{c}{\begin{tabular}[c]{@{}c@{}}    Proposed \end{tabular}} &   PDM&   \textbf{12.21}&   \textbf{22.44}&   8.46&   \textbf{10.48}\\ \hline
\end{tabular}

\label{tab:S5}
\end{table}
In addition, we perform a parameter-efficient validation experiment (see Table~\ref{tab:parameter}), which demonstrates that our MR loss achieves competitive performance with fewer parameters and less inference time. We also compare our method against pure diffusion-based synthesis in Table~\ref{tab:S5}, where our approach yields superior synthesis quality. Finally, we validate the necessity of the PDM: by generating initial images with PDM and refining them with the subsequent refinement network, our overall strategy proves to be optimal.

\begin{table}[htbp]
\footnotesize
\vspace{-1mm}
    \centering
    \caption{
    The ablation study showed PDM boosted output quality versus baseline refinement network.
    %  fMRI synthesis achieved +1.54 dB PSNR and +4.12\% SSIM improvements. 
    % dMRI showed smaller gains (+0.12 dB PSNR/+0.92\% SSIM). 
    fMRI's marked improvements may enhance diagnostic performance (S1 vs. S4 in Table \ref{tab:diagnosis}).
    }
    \vspace{-1mm}
    \begin{tabular}{cccccc}
    \hline
    \multirow{2}{*}{\makecell[c]{Refine \\ Network}} & \multirow{2}{*}{PDM} & \multicolumn{2}{c}{fMRI} & \multicolumn{2}{c}{dMRI} \\
    \cline{3 - 6} 
     &  & PSNR & SSIM & PSNR& SSIM\\
    \hline
     $\checkmark$&  $\times$& 28.29 & 86.72 & 29.98 & 76.35 \\
    \hline
     $\checkmark$ &   $\checkmark$ &     \textbf{29.83} &   \textbf{90.84} &   \textbf{30.10} &   \textbf{77.29} \\
    \hline
    \end{tabular}
    \label{tab:S4}
    \vspace{-2mm}
\end{table}

\begin{table}[htbp]
\footnotesize
\vspace{-2mm}
    \centering
    \caption{The pattern-aware module demonstrated consist improvements across  metrics.
    % : +3.77\% accuracy , +3.8\% precision , +6.98\% recall for  MCI detection, and +6.95\% AUC.
    }
    \vspace{-1mm}
    \begin{tabular}{c|ccccc}
        \hline
        & \multicolumn{4}{c}{Metrics} \\ \hline
        Pattern-aware & Acc & Pre & Rec & F1-score & AUC \\ \hline
        $\times$ & 73.58 & 44.27 & 41.74 & 41.87 & 58.70 \\
        $\checkmark$ & \textbf{77.35} & \textbf{48.07} & \textbf{48.82} & \textbf{48.44} & \textbf{65.65} \\ \hline
    \end{tabular}
    \label{tab:S3}
    \vspace{-5mm}
\end{table}

%{-6mm}
\section*{Conclusion}

In this paper, we propose \textbf{\textit{PDS}}, a novel framework for bidirectional 3D fMRI-dMRI synthesis that integrates disease semantics, tissue preservation, and microstructure refinement. Our disease-semantic-informed architecture captures pathological \textbf{patterns} across modalities through a pattern-aware module. Cascaded refinement modules further enhance structural consistency and biological fidelity via tissue projection and microstructural perception. Extensive experiments on three datasets demonstrate state-of-the-art performance, with significant gains in PSNR (+1.54 dB fMRI, +1.02 dB dMRI) and SSIM (+4.12\%, +2.2\%). Clinically, synthetic data achieved 67.92\% diagnostic accuracy using hybrid real-synthetic inputs. This work provides a foundation for cost-effective, multimodal neuroimaging with potential applications in early detection, longitudinal monitoring, and personalized treatment of neurodegenerative diseases.

\section*{References}

\bibliography{refs}

\end{document}